\pdfoutput=1

\documentclass[11pt]{article}

\usepackage[final]{acl}

\usepackage{times}
\usepackage{latexsym}

\usepackage[T1]{fontenc}

\usepackage[utf8]{inputenc}

\usepackage{microtype}

\usepackage{inconsolata}

\usepackage{graphicx}
\usepackage{makecell}

\usepackage{booktabs}

\usepackage{amsmath}

\usepackage{wrapfig}

\usepackage{multirow}

\usepackage{colortbl}

\usepackage{tablefootnote}

\usepackage[stable]{footmisc}

\usepackage{amssymb}

\usepackage{pifont}

\usepackage{xcolor}
\usepackage{tcolorbox}
\usepackage{soul}
\usepackage{float}

\definecolor{lightgreen}{rgb}{0.6, 1, 0.6}
\definecolor{lightred}{rgb}{1, 0.6, 0.6}
\definecolor{lightyellow}{rgb}{1, 1, 0.6}
\definecolor{lightblue}{rgb}{0.6, 0.8, 1}
\definecolor{lightgrey}{rgb}{0.93, 0.93, 0.93}

\newcommand{\hlgreen}[1]{{\sethlcolor{lightgreen}\hl{#1}}}
\newcommand{\hlred}[1]{{\sethlcolor{lightred}\hl{#1}}}

\newcommand{\hlgrey}[1]{{\sethlcolor{lightgrey}\hl{#1}}}

\title{LongRAG: A Dual-Perspective Retrieval-Augmented Generation Paradigm for Long-Context Question Answering}

\author{
 \textbf{Qingfei Zhao\textsuperscript{1,2,\footnotemark[2]}},
 \textbf{Ruobing Wang\textsuperscript{1,2}},
 \textbf{Yukuo Cen\textsuperscript{4}},\\
 \textbf{Daren Zha\textsuperscript{1}},
 \textbf{Shicheng Tan\textsuperscript{3}},
 \textbf{Yuxiao Dong\textsuperscript{3}},
 \textbf{Jie Tang\textsuperscript{3,\thanks{Corresponding author}}}
\\
 \textsuperscript{1}Institute of Information Engineering,
Chinese Academy of Sciences;\\
 \textsuperscript{2}School of Cyber Security,
University of Chinese Academy of Sciences;\\
 \textsuperscript{3}Tsinghua University;
 \textsuperscript{4}Zhipu AI\\
 \texttt{\{zhaoqingfei, wangruobing, zhadaren\}@iie.ac.cn}, \texttt{yukuo.cen@zhipuai.cn}\\
 \texttt{tsctan@foxmail.com}, \texttt{\{yuxiaod, jietang\}@tsinghua.edu.cn}
}

\begin{document}
\maketitle
\renewcommand{\thefootnote}{\fnsymbol{footnote}}
\footnotetext[2]{Work done when QZ interned at Zhipu AI}
\renewcommand{\thefootnote}{\arabic{footnote}}
\begin{abstract}
Long-Context Question Answering (LCQA), a challenging task, aims to reason over long-context documents to yield accurate answers to questions.
Existing long-context Large Language Models (LLMs) for LCQA often struggle with the "\textit{lost in the middle}" issue.
Retrieval-Augmented Generation (RAG) mitigates this issue by providing external factual evidence. However, its chunking strategy disrupts the global long-context information, and its low-quality retrieval in long contexts hinders LLMs from identifying effective factual details due to substantial noise.
To this end, we propose LongRAG, a general,
dual-perspective, and robust LLM-based RAG system paradigm for LCQA to enhance RAG's understanding of complex long-context knowledge (i.e., global information and factual details). We design LongRAG as a plug-and-play paradigm, facilitating adaptation to various domains and LLMs.
Extensive experiments on three multi-hop datasets demonstrate that LongRAG significantly outperforms long-context LLMs (up by 6.94\%), advanced RAG (up by 6.16\%), and Vanilla RAG (up by 17.25\%).
Furthermore, we conduct quantitative ablation studies and multi-dimensional analyses, highlighting the effectiveness of the system's components and fine-tuning strategies. Data and code are available at \url{https://github.com/QingFei1/LongRAG}.
\end{abstract}
\section{Introduction}
Large language models (LLMs), such as GPT \citep{DBLP:conf/nips/BrownMRSKDNSSAA20}, 
GLM \citep{DBLP:journals/corr/abs-2210-02414}
and LLaMA \citep{DBLP:journals/corr/abs-2302-13971}, boost the real-world development of multiple scenarios.
Long-context question answering (LCQA) \citep{DBLP:conf/naacl/CaciularuDGC22}, which has been recently advanced significantly by LLMs, is a complex task that requires reasoning over a long document or multiple documents to 
\begin{figure}[t]
  \includegraphics[width=0.95\columnwidth]{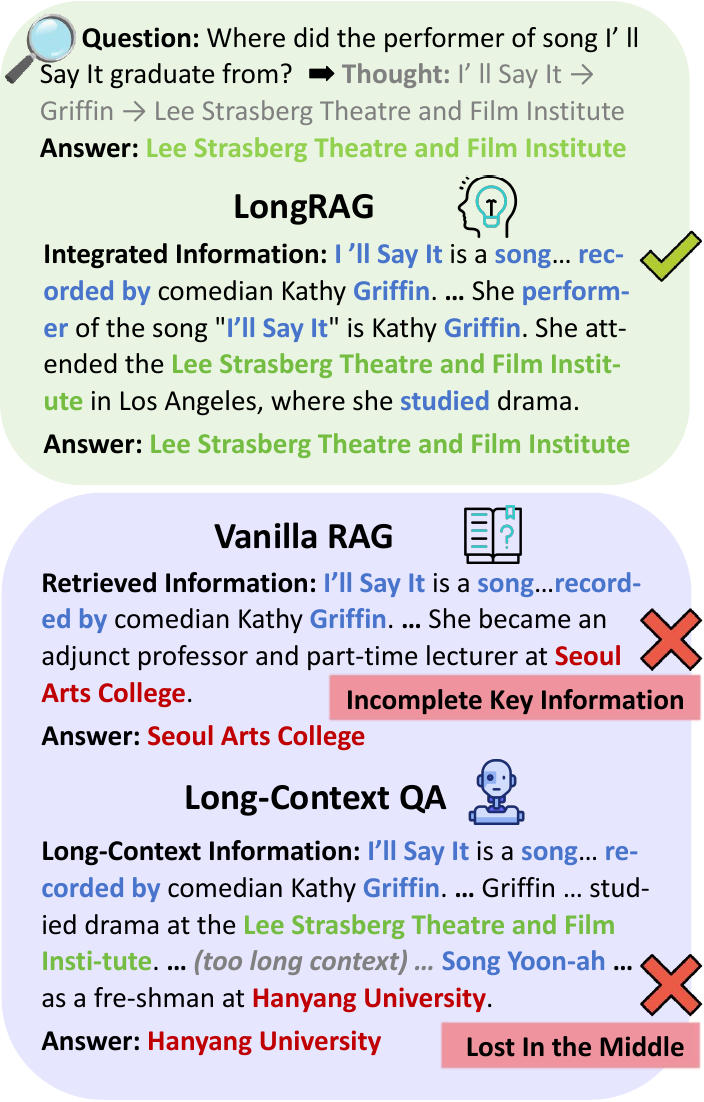}
  \caption{\textbf{Examples of Different Methods.} Long-Context LLMs and Vanilla RAG face "\textit{lost in the middle}" and "\textit{incomplete key information}" issues, while LongRAG addresses them, yielding a perfect answer.}
  \label{fig:1_task}
\end{figure}
provide accurate answers to questions.
Recently, several long-context LLMs have been introduced, such as Gemini~\citep{DBLP:journals/corr/abs-2312-11805} and GPT-4-128k, capable of ingesting entire relevant documents and generating answers directly. 
However, as shown in Figure~\ref{fig:1_task}, they frequently encounter the “\textit{lost in the middle}” issue~\citep{DBLP:journals/tacl/LiuLHPBPL24}, that is, when the relevant context is in the middle of the document (rather than the beginning and end), they are prone to sub-optimal or even incorrect responses.
Instead, the Retrieval-Augmented Generation (RAG) system \citep{DBLP:journals/corr/abs-2312-10997,DBLP:conf/icml/GuuLTPC20} offers an alternative approach, mitigating this issue by employing a fixed-length chunking strategy~\citep{llamaindex}. This strategy ensures the input to the LLM is concise and highly relevant to the question.

Nevertheless, Vanilla RAG remains insufficient for the LCQA task due to two major limitations.
\textbf{First}, the chunking strategy disrupts the contextual structure and background information in long documents (global information).
Some chunks may contain incomplete information~\citep{DBLP:journals/corr/abs-2302-14502},
thereby causing LLMs to draw upon irrelevant context or fall back on their internal parameterized knowledge, potentially leading to inaccurate responses.
\textbf{As depicted in Figure~\ref{fig:1_task}}, Vanilla RAG only retrieves "\textit{Griffin}" as the performer of "\textit{I'll say it}" but misses the university from which "\textit{Griffin}" graduated.
Although the "\textit{university}" is mentioned in the same paragraph,
the system ultimately produces an incorrect response.
\textbf{Second}, low evidence density in long-context documents can lead to low retrieval quality.
Considerable noise present in long-context documents impairs LLMs' capacity to accurately identify key information (factual details), resulting in the retrieval of low-quality chunks and ultimately leading to erroneous answers
(\citealp{DBLP:journals/corr/abs-2309-01219}; \citealp{DBLP:conf/aaai/0011LH024}).
Recently, several advanced RAG systems have attempted to mitigate the aforementioned issues.
Specifically,
Self-RAG~\citep{DBLP:journals/corr/abs-2310-11511} employs self-reflection tokens to facilitate the autonomous exploration of global information in a corpus. However, its reliance on the accuracy of reflection tokens may result in the potential deletion of valid retrieval chunks with factual details.
CRAG~\citep{DBLP:journals/corr/abs-2401-15884} evaluates the question relevance of each chunk individually to enhance the identification of factual details. Nevertheless, it overlooks the connections between chunks, provoking low-quality evaluation when valid details span multiple chunks, potentially leading to the omission of crucial factual details.

In our work, we propose LongRAG, a general, dual-perspective, and robust RAG system paradigm that effectively addresses the above-mentioned issues for LCQA, comprising four plug-and-play components with multiple strategies: a hybrid retriever, an LLM-augmented information extractor, a CoT-guided filter, and an LLM-augmented generator.
LongRAG enhances the RAG system's ability to mine global long-context information and identify factual details.
Specifically, the long-context extractor employs a mapping strategy to orderly extend the semantic space of retrieved chunks into a higher dimensional long-context semantic space, then refining global information and contextual structure among chunks.
Meanwhile, the CoT-guided filter utilizes the Chain of Thought (CoT)~\citep{DBLP:conf/nips/Wei0SBIXCLZ22} to provide global clues according to the knowledge of all retrieved chunks, instructing LLMs to carefully review factual details and precisely filter out irrelevant chunks. This improves evidence density and enhances RAG's ability to understand complex and lengthy contexts.
Additionally, we have curated an automated instruction data pipeline for constructing a high-quality dataset for fine-tuning. This fine-tuning strategy significantly enhances the “\textit{instruction-following}” capabilities of the system's core components.
It is also convenient to transfer LongRAG to other domains by leveraging the pipeline and fine-tuning strategy.

Extensive performance comparisons and quantitative ablation studies conducted on three multi-hop datasets from LongBench~\citep{bai2023longbench} demonstrate the superiority and effectiveness of LongRAG.
The results suggest that LongRAG significantly outperformed both long-context LLMs and advanced RAG methods.
We also discuss LongRAG's performance with different fine-tuned LLMs and confirm its strong robustness and transferability.
To sum up, our contributions are summarized as follows:
\textbf{1)} We construct LongRAG, a general, dual-perspective, and robust RAG system paradigm. It significantly surpasses long-context LLM (up by 6.94\%), mainstream advanced RAG (up by 6.16\%), and Vanilla RAG (up by 17.25\%).
\textbf{2)} We identify and address RAG's limitations in LCQA. We develop two plug-and-play components (i.e., Information Extractor and CoT-guided Filter) to explore global information and factual details, enhancing understanding of complex long contexts.
\textbf{3)} We implement a novel automated fine-tuning data construction pipeline and a multi-task training strategy with multi-length long-context data. They facilitate the application of our paradigm to diverse specific-domain data in real-world scenarios.

\section{Related Works}
\subsection{Long-Context LLMs}
LLMs usually need to handle complex and long-context inputs in the real world.
The context window length of LLMs is limited by their training sequence length, and inputs exceeding this window may result in considerable performance degradation (\citealp{DBLP:journals/corr/abs-2303-18223}; \citealp{DBLP:journals/corr/abs-2401-01325}).
Thus, recent studies focus on scaling the limited context length of existing LLMs to accommodate tasks requiring long contexts, e.g., long-context question-answering.  
Methods for scaling the context length are categorized into two main types: 1) One is methods for training or fine-tuning with long contexts, such as
RMT~\citep{DBLP:conf/nips/BulatovKB22}, Position Interpolation~\citep{DBLP:journals/corr/abs-2306-15595}, YaRN~\citep{DBLP:journals/corr/abs-2309-00071}, Activation Beacon~\citep{DBLP:journals/corr/abs-2401-03462}, LongLoRA~\citep{DBLP:journals/corr/abs-2309-12307}, LongRoPE~\citep{DBLP:journals/corr/abs-2402-13753}, and LongAlign~\citep{DBLP:journals/corr/abs-2401-18058}; 2) the other is non-fine-tuned methods include restricted attention-based approaches (\citealp{DBLP:journals/corr/abs-2308-16137}; \citealp{DBLP:journals/corr/abs-2309-17453}; \citealp{DBLP:journals/corr/abs-2402-10685})
and context compression methods (\citealp{DBLP:journals/corr/abs-2310-06839}; \citealp{DBLP:conf/emnlp/0001DGL23}).
Generally, non-fine-tuned methods allow for plug-and-play and low-cost scaling LLMs. Fine-tuned methods typically show better performance but require higher training and data costs.
\subsection{Retrieval-Augmented Generation}
With the advent of the GPT
era,
RAG \cite{lewis2020retrieval,DBLP:conf/icml/GuuLTPC20} is regarded as a powerful technology for improving the response quality of LLMs \citep{DBLP:conf/iclr/IzacardG21,DBLP:journals/corr/abs-2210-11416}.
RAG alleviates issues such as outdated and long-tail knowledge (\citealp{DBLP:journals/corr/abs-2301-00303}; \citealp{DBLP:conf/icml/KandpalDRWR23}), hallucinations (\citealp{DBLP:conf/cikm/ChenFYWFL0LX23}; \citealp{DBLP:conf/sigir-ap/ZucconKS23}), and lack of domain expertise (\citealp{DBLP:conf/emnlp/LiCZPMLS23}; \citealp{DBLP:journals/corr/abs-2304-08979}) of LLMs by leveraging external knowledge, i.e., Wikipedia. 
Despite the success of RAG, its chunking strategy and direct incorporation of retrieved chunks into the generator result in incomplete information and substantial noise. 
Recently, advanced RAG models have been proposed to address these issues by filtering or re-ranking the retrieved knowledge to reduce noise (\citealp{DBLP:journals/corr/abs-2310-01558}; \citealp{DBLP:journals/corr/abs-2401-15884}; \citealp{DBLP:conf/emnlp/Zhuang0KZ23}), designing a chunk-free strategy to mitigate semantic loss (\citealp{DBLP:journals/corr/abs-2402-09760}), and employing active retrieval to mine information (\citealp{DBLP:journals/corr/abs-2310-11511}; \citealp{DBLP:conf/emnlp/JiangXGSLDYCN23}).

\subsection{Domain-Specific Fine-Tuning for RAG}
Fine-tuning has gradually become a popular strategy \citep{DBLP:journals/corr/abs-2401-06954} for enhancing the capabilities of components of RAG.
Existing works include fine-tuning retrieval-related components to achieve better retrieval outcomes \citep{DBLP:journals/corr/abs-2401-15884}, fine-tuning generators for more personalized outputs \citep{DBLP:journals/corr/abs-2403-10131}, and employing collaborative fine-tuning \citep{DBLP:journals/corr/abs-2310-01352}. Additionally, \citet{DBLP:conf/nips/ZhouLX0SMMEYYZG23} discovered that fine-tuning LLMs with a limited quantity of high-quality data significantly enhances the performance of LLMs. This finding provides a robust theoretical basis for collaboratively fine-tuning multiple components within advanced RAG methodologies at a minimal data expense.
\begin{figure*}[t]
  \includegraphics[width=16cm]{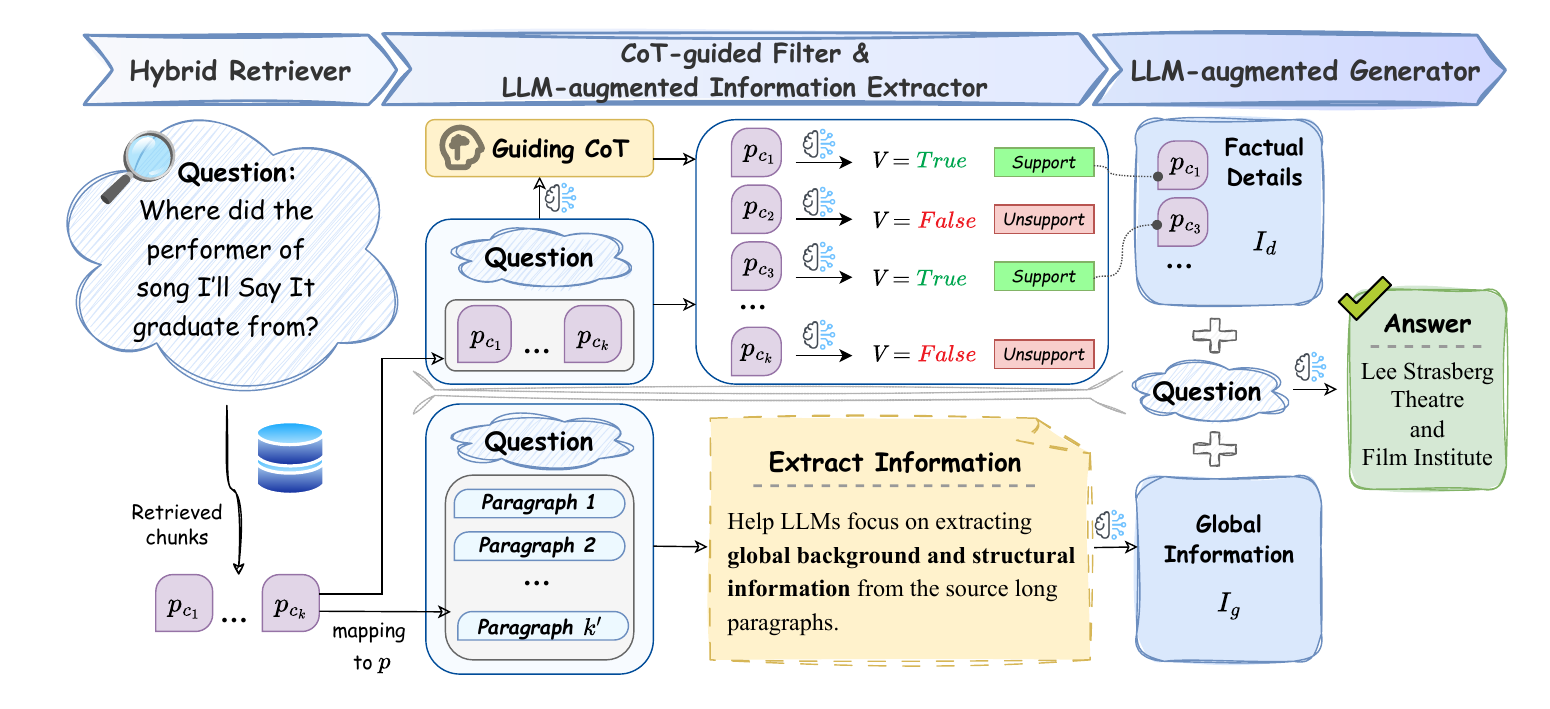}
  \caption{\textbf{An overview of LongRAG.} Our system involves four sub-components: Hybrid Retriever receives a question and retrieves the top-$k$ most relevant chunks $p_c$; CoT-guided Filter generates global key clues to analyze their relevance one by one, obtaining a set of "\texttt{True}" chunks as $I_d$; Meanwhile, LLM-augmented Information Extractor sequentially maps $p_c$ to the source long-context paragraph $p$ to extract effective global information $I_g$; LLM-augmented Generator promotes knowledge interaction between $I_g$ and $I_d$ to generate the final answer.
  }
  \label{fig:2_overall}
\end{figure*}

\section{Preliminaries}
\subsection{Task Definition}
Following the structure of Vanilla RAG (a retriever $\mathcal{R}$ and a generator $\mathcal{G}$), the LongRAG system (\textit{cf.,} Figure~\ref{fig:2_overall}) includes a \textbf{Long-Context Extractor} $\mathcal{E}$ and a \textbf{CoT-guided Filter} $\mathcal{F}$ after retrieval to extract global information $I_g$ and identify factual details $I_d$. 
Specifically, given a question $q\in Q$ and a long-context corpus $\mathcal{C}$, $\mathcal{R}$ receives a $q$ and retrieves the top-$k$ most relevant chunks $p_c \in P_c$. 
These $p_c$ are obtained by segmenting source paragraphs $p \in P $.
We then input $p$ into $\mathcal{E}$, obtaining $I_g$, and $p_c$ into $\mathcal{F}$ to identify chunks containing factual details, defined as $I_d$, which are subsequently used by $\mathcal{G}$ to generate a final answer to the question.
It is worth noting that when discussing the system, $P$ represents the source long-context paragraphs mapping from retrieved chunks $P_c$. However, when discussing fine-tuning instruction data $\mathcal{D}$, $P$ denotes all corresponding paragraphs given for a question, including predefined supporting paragraphs $P_s$ and given distracting paragraphs $P_d$.
\subsection{Fine-Tuning Data Construction}
To improve the "\textit{instruction following}" ability of components and learn long-context styles,
we craft a small but high-quality instruction-following dataset for supervised fine-tuning (SFT), named \textbf{LRGinstruction}, via ChatGLM3-32B-128k~\citep{DBLP:conf/acl/DuQLDQY022,DBLP:conf/iclr/ZengLDWL0YXZXTM23} as teacher LLM. 
We select the training sets of three complex English multi-hop datasets released by~\citet{DBLP:conf/acl/TrivediBKS23} -- HotpotQA \citep{DBLP:conf/emnlp/Yang0ZBCSM18}, 2WikiMultiHopQA\citep{DBLP:conf/coling/HoNSA20}, and MusiQue \citep{DBLP:journals/tacl/TrivediBKS22}, as well as the English dataset QASPER with longer contexts
\citep{DBLP:conf/naacl/DasigiLBCSG21}
, to jointly develop our LRGinstruction. Among them, QASPER with more lengthy contexts promotes LLMs to further learn the long-context style.
The construction pipeline is automated, that is, you can automatically generate high-quality fine-tuning instruction data from any specific domain. In addition, the results of experiments indicate that we only need 2600 samples to fine-tune the LLMs used in components to achieve good performance in LCQA tasks.
The construction pipeline is introduced as follows (\textbf{more details in Appendix~\ref{Details of LRGinstruction}}).

\noindent\textbf{Data Pre-Processing.}
To learn long-context style, we discard any question-answer pairs with insufficient context length (see details in Appendix~\ref{Appendix: Data Pre-Processing}).
Then, we keep all supporting paragraphs of questions $P_{s}$ and randomly retain a subset of distracting paragraphs $P_{d}$. The random strategy is designed to simulate the distribution of the number of recalls executed in reality.
To sum up, we define the elements of pre-processed dataset as follows: question $q\in Q$, multiple corresponding paragraphs $p\in P$, including supporting paragraphs $P_{s}$ and distracting paragraphs $P_{d}$ to the question, and answer $\alpha \in A$, mathematically $\left \langle Q, \left \{P_{s}  \cup P_{d}  \right \}, A \right \rangle$.

\noindent\textbf{Long-Context Extractor Data.}
We fine-tune the long-context extractor to improve its capacity to extract global information from the source long paragraphs.
First, we consider all $P_{s}$ of each question as effective global information. These questions and their global information serve as input for zero-shot in-context learning (ICL) to gain global background and structure information, which act as golden outputs (see Appendix~\ref{Appendix: Long-Context Extractor Data} for details).
Subsequently, to enhance the robustness of the pipeline, we validate the efficacy of the golden outputs via an LLM-based self-evaluator and retain the golden outputs that are deemed valid.

\noindent\textbf{CoT-guiding Data \& Filtering Data.}
The training data for the CoT-guided filter component is constructed in two stages: the CoT guidance and the filtering stage.
Key insights and clues for question resolution reside within $P_{s}$. Thus, for the CoT guidance stage, the LLM is expected to examine the semantic relations and factual details for question-solving within $P_{s}$ to generate a guiding CoT. This process also employs a self-evaluator to evaluate the reliability of the CoT outputs as golden data. In the subsequent filtering stage, 
We merge $q$ with a corresponding $p$ and its guiding CoT as the gold data (see Appendix~\ref{Appendix: CoT-guiding Data and Filtering Data} for details). $P_s$ and $P_d$ each account for half in $P$.

\noindent\textbf{Task-Oriented Data.}
Question-answer pairs $\left \langle Q, A \right \rangle$ and $P$ are already present in $\mathcal{D}$, and we simply need to reorganize their format.

\section{The LongRAG System}
\subsection{Hybrid Retriever}
The hybrid retriever begins with a given question and then recalls $k$ chunks.
Before the retrieval, the long-context $p$ requires further segmentation into chunks $p_{c}$.
Specifically, we impose a length limit on chunks, with sentences as the smallest division unit. 
We then employ a sliding window
to extend the context by adding overlapping content from the end of the previous sentence, 
preventing semantic disruption at truncation points.
Short chunks at the end of $p$ are merged with preceding chunks to ensure better semantic cohesion.
Inspired by Re$^2$G~\citep{DBLP:journals/corr/abs-2207-06300}, we utilize a dual-encoder\footnotemark[1] structure for rapid retrieval at a coarse-grained level, and a cross-encoder\footnotemark[2]
\footnotetext[1]{We use E5-large model for dual-encoder:\url{https://huggingface.co/intfloat/multilingual-e5-large}}\footnotetext[2]{We use mMiniLM as cross-encoder model: \url{https://huggingface.co/nreimers/mmarco-mMiniLMv2-L12-H384-v1}}to capture the deep semantic interaction for further retrieval at a fine-grained level.
The engineering implementation ensures efficient retrieval through the use of FAISS~\citep{johnson2019billion}.
\subsection{LLM-augmented Information Extractor}
\label{LLM-based information extractor}
In long-context QA with low evidence density, the complete evidence supporting answers is usually scattered across multiple locations. From a global perspective, this evidence not only contains its own knowledge but also implicitly stores logical and sequential connections among chunks. Retrieved chunks, truncated by fixed windows, struggle to carry additional global information.
Furthermore, 
when the retrieved chunks originate from the same $p$, their order may be inconsistent with the original semantic order in $p$, resulting in providing disordered semantic information to downstream LLMs.
To address these issues, we map the short-form chunks $p_{c}$ back to their source long-context paragraphs $p$,
using a mapping function $f_{m}(\cdot)$:
\begin{equation}
    f_{m}(p_{c_{1}}, p_{c_{2}},\cdots , p_{c_{k}}) \rightarrow {p_{1}, p_{2},\cdots , p_{k'}}
\end{equation}
where $k$ and $k'$ ($k\leq k'$) denote the number of pre-mapping $p_c$ and post-mapping $p$, respectively. When multiple $p_c$ correspond to the same $p$, we keep only the $p$ corresponding to the $p_c$ with the highest semantic similarity to the question $q$.
This mapping strategy maximizes the recovery of the context of question-relevant source paragraphs. Then, we concatenate $k'$ paragraphs $p$ and feed them into the prompt (see Appendix~\ref{Prompts of LongRAG system}) of the LLM-augmented information extractor for employing zero-shot ICL.
\begin{equation}
    I_{g}=\text{LLM}(prompt_{e}(q, p_{1} || p_{2}|| \cdots ||p_{k'}))
\end{equation}
The prompt template of the LLM-augmented information extractor, defined as $prompt_{e}(\cdot)$, guides the LLM to ultimately obtaining global information $I_g$ enriched with extensive long-context background and structural knowledge.

\subsection{CoT-guided Filter}
\label{CoT-guided filter}
It is not always the case that retrieved chunks $p_{c}$ will assist in answering questions, particularly in multi-hop questions that involve complex reasoning chains and long-context paragraphs with low evidence density.
The retrieved chunks usually contain substantial redundancy; some of chunks can even be entirely redundant.
This complexity makes it difficult to ascertain whether a chunk holds the key information for solving multi-hop questions.

To address this, we develop the CoT-guided filter with a two-stage strategy.
The initial stage, CoT guidance, generates a CoT with a global perspective based on the retrieval semantic space, outlining the global clues for answering the question.
Here's the mathematical expression of CoT-guidance stage:
\begin{equation}
    CoT=\text{LLM}(prompt_{c}(q, p_{c_1}|| \cdots ||p_{c_{k}}))
    \label{4.3:CoT}
\end{equation}
where $k$ denotes the number of chunks $p_c$, and $prompt_{c}(\cdot)$ is the prompt template of yielding CoT based on LLMs.
Subsequently, in the filtering stage, these CoTs serve as global clues, guiding LLMs step by step to focus on relevant knowledge throughout the reasoning chain.
They equip filters with the ability to judge the relevance between questions and chunks using a high-dimensional perspective.
This aids the system in inferring multi-hop semantic associations and meticulously examining all available factual details in contexts of low evidence density.
Overall, this phase achieves high-quality identification of factual details and secures reliable relevance labels for question-chunk pairs.
We use these labels to precisely filter irrelevant chunks $p_c$ and avoid deleting crucial factual details, thus ensuring low redundancy input for the downstream generator.
\begin{equation}
\begin{aligned}
V(q, p_c, \text{CoT}) = 
\begin{cases} 
\text{True,} & \text{if }\text{<support>} \\
\text{False,} & \text{otherwise}
\end{cases} \\
I_{d} = \{ p_c \mid V(q, p_c, \text{CoT}) = \text{True} \}
\end{aligned}
\label{4.3:Filter}
\end{equation}
Equation (\ref{4.3:Filter}) describes the process of the filtering stage. $V(\cdot)$ returns a binary label to assess whether the chunk $p_{c}$ supports answering the $q$ according to the clues within the CoT.
We iteratively assess each $p_{c}$ via the function $V(\cdot)$. 
These chunks marked as "True" are considered as a set of chunks containing factual details information, defined as $I_d$.

\begin{table*}[t]
  \centering
    \fontsize{8}{9}\selectfont
    \setlength{\tabcolsep}{9pt}
    \begin{tabular}{ccccc}
    \toprule
    Model & \textbf{HotpotQA} & \textbf{2WikiMQA} & \textbf{MusiQue} & Average \\
    \midrule
    \rowcolor{gray!20}
    \multicolumn{5}{c}{\# Long-Context LLM Methods \#} \\
    LongAlign-7B-64k (\textit{Llama2})~\citep{DBLP:journals/corr/abs-2401-18058} & \textbf{48.85} & 28.56 & 25.14 & 34.18\\
    LongLoRA-13B-32k (\textit{Llama2})~\citep{DBLP:journals/corr/abs-2309-12307} & 47.45 & \textbf{42.92} & \textbf{29.46} & \textbf{39.94} \\
    \rowcolor{gray!20}
    \multicolumn{5}{c}{\# Advanced RAG Methods \#} \\
    CFIC-7B (\textit{Llama2})~\citep{DBLP:journals/corr/abs-2402-09760} & 34.00  & - & 14.70  & 24.35 \\
    CRAG (\textit{GPT-3.5-Turbo})~\citep{DBLP:journals/corr/abs-2401-15884} & \textbf{52.04}  & 41.13  & \textbf{25.34}  & 39.50 \\
    Self-RAG (\textit{GPT-3.5-Turbo})~\citep{DBLP:journals/corr/abs-2310-11511} & 50.51  & \textbf{46.75}  & 24.62  & \textbf{40.63} \\
    \rowcolor{gray!20}
    \multicolumn{5}{c}{\# RAG-Base (Vanilla RAG) \#} \\
    Vicuna-v1.5-7B-16k~\citep{DBLP:conf/nips/ZhengC00WZL0LXZ23} & 38.63  & 27.92 & 15.68  & 27.41 \\
    Qwen-1.5-7B-32k~\citep{DBLP:journals/corr/abs-2309-16609} & 45.70  & 34.69  & 25.08  & 35.16 \\
    Llama3-8B-8k~\citep{DBLP:journals/corr/abs-2302-13971} & 48.25  & 43.47  & 19.66  & 37.13 \\
    ChatGLM3-6B-32k~\citep{DBLP:conf/acl/DuQLDQY022} & 52.57  & 42.56  & 25.51  & 40.21 \\
    GPT-3.5-Turbo-16k & 50.17  & 45.32  & 21.84  & 39.11 \\
    GPT-3.5-Turbo & 52.31  & 43.44  & 25.22  & 40.32 \\
    Llama3-70B-8k & 52.33 & 50.23 & 25.49 & 42.68 \\
    GLM-4 & \textbf{57.41}  & \textbf{52.91}  & \textbf{27.55}  & \textbf{45.96} \\
    \rowcolor{gray!20}
    \multicolumn{5}{c}{\# \textbf{Ours} with SFT \#} \\
    LongRAG-Llama2-7B-4k & 53.85 & 45.61 & 26.22 & 41.89 \\
    LongRAG-Llama2-13B-4k & \textbf{57.05} & 49.95 & \textbf{33.63} & 46.88 \\
    LongRAG-Qwen-1.5-7B-32k & 52.91\,\,(7.21↑) & 46.65\,\,(11.96↑) & 31.85\,\,(6.77↑) & 43.80\,\,(8.65↑) \\
    LongRAG-Llama3-8B-8k & 52.39\,\,(4.14↑) & 49.67\,\,(6.20↑) & 31.70\,\,(12.04↑) & 44.59\,\,(7.46↑) \\
    LongRAG-Vicuna-v1.5-7B-16k & 55.55\,\,\underline{(16.92↑)} & 50.13\,\,\underline{(22.21↑)} & 28.29\,\,\underline{(12.61↑)} & 44.66\,\,\underline{(17.25↑)} \\
    LongRAG-ChatGLM3-6B-32k & 55.93\,\,(3.36↑) & \textbf{54.85}\,\,(12.29↑) & 33.00\,\,(7.49↑) & \textbf{47.93}\,\,(7.71↑) \\
    \rowcolor{gray!20}
    \multicolumn{5}{c}{\# \textbf{Ours} without SFT \#} \\
    LongRAG-GPT-3.5-Turbo & 56.17\,\,(3.86↑) & 51.37\,\,\underline{(7.93↑)} & 32.83\,\,(7.61↑) & 46.79\,\,(6.47↑) \\
    LongRAG-GPT-3.5-Turbo-16k & 59.11\,\,\underline{(8.94↑)} & 51.25\,\,(5.93↑) & 30.37\,\,(8.53↑) & 46.91\,\,\underline{(7.80↑)} \\
    LongRAG-GLM-4 & \textbf{62.11}\,\,(4.70↑) & \textbf{57.16}\,\,(4.25↑) & \textbf{38.40}\,\,\underline{(10.85↑)} & \textbf{52.56}\,\,(6.60↑) \\
    \bottomrule
    \end{tabular}%
    \caption{\textbf{Results (\%) of overall performance on three multi-hop datasets.} The "\hlgrey{Grey Areas}" represent different categories of baselines or our system with different fine-tuning settings. “\textbf{Bold Font}” denotes the highest absolute value, while "\underline{Underlined Font}" expresses the highest relative gain value compared to Vanilla RAG. Ours with (or without) SFT indicates we employ fine-tuned (or non-fine-tuned) LLMs in all LLM-augmented components. All model types are "chat". We calculate the increase in ours compared to Vanilla RAG, such as "17.25↑".}
  \label{tab:overall}%
\end{table*}%

\subsection{LLM-augmented Generator}
\label{LLM-based generator}
Global information $I_{g}$ encompasses both background and structural information within the long-context corpus, while factual details information $I_{d}$ refers to the filtered chunk set with minimal noise and crucial evidence details. The generator boosts the interaction of knowledge across these two perspectives to produce answers $\alpha $ to questions.
Here is the formula for the generator $\mathcal{G}$, where $prompt_{g}(\cdot)$ is the prompt template of generator:
\begin{equation}
    \alpha =\text{LLM}(prompt_{g}(I_{g}, I_{d}))
    \label{4.4:Gen}
\end{equation}

\subsection{Instruction-Tuning}
We adopt a collection of industry-leading models as our foundational LLMs: ChatGLM~\citep{DBLP:conf/acl/DuQLDQY022,DBLP:journals/corr/abs-2210-02414}, Qwen1.5~\citep{DBLP:journals/corr/abs-2309-16609}, Vicuna~\citep{DBLP:conf/nips/ZhengC00WZL0LXZ23}, Llama2, and Llama3~\cite{DBLP:journals/corr/abs-2302-13971}.
They are all open-source and support multi-lingual, multi-tasking.
We have fine-tuned them using 2,600 high-quality data sourced from LRGinstruction.
Specifically, we employ all four types of data in LRGinstruction collectively to train a model that is used in the extractor, the filter, and the generator.
Furthermore, this data has undergone length filtering and has been standardized into a QA instruction style.
During training, all models utilize the Llama-factory library and 8xA100 GPUs (80G each), employing training methods with DeepSpeed+ZeRO3+CPU offloading+flash attention strategies \citep{DBLP:conf/kdd/RasleyRRH20,DBLP:conf/nips/DaoFERR22}.
The training parameters are set with a batch size of 8, a gradient accumulation step of 12, and 3 epochs (totaling 81 steps).

\section{Experiment}
\subsection{Experimental Setup}
\textbf{Datasets \& Evaluation.}
We select three challenging multi-hop datasets -- HotpotQA, 2WikiMultiHopQA (2WikiMQA), and MusiQue -- from the Longbench \citep{bai2023longbench} for evaluation, rather than using raw datasets.
We standardize these data to adapt to RAG tasks \textbf{(more details in Appendix~\ref{Details of Experimental Datasets})}, and report the F1-score as evaluation metrics for all three datasets. Statistics of experimental datasets are shown in Table~\ref{tab1:dataset}.
\begin{table}[t]
  \centering
    \setlength{\tabcolsep}{5pt}
    \fontsize{8}{9}\selectfont
    \begin{tabular}{cccc}
    \toprule
    \textbf{Dataset} & \textbf{HotpotQA} & \textbf{2WikiMQA} & \textbf{MuSiQue} \\
    \midrule
    Num of Samples & 200  & 200  & 200  \\
    Avg. Length of $p$ & 1092  & 535  & 1032  \\
    Num of $p$ & 1715  & 1464  & 1877  \\
    Avg. Length of $P$ & 9151  & 4887  & 11214  \\
    \bottomrule
    \end{tabular}%
    \caption{\textbf{Statistics of experimental data}. "Avg. Length" stands for the average word count.}
  \label{tab1:dataset}%
\end{table}%

\noindent\textbf{Baselines \& LLMs.}
To validate the superiority of our LongRAG in multiple dimensions, we utilize three categories of baselines:
1) Long-Context LLM Methods -- \textbf{LongAlign}~\citep{DBLP:journals/corr/abs-2401-18058} and \textbf{LongLoRA}~\citep{DBLP:journals/corr/abs-2309-12307}; 2) Advanced RAG Methods -- \textbf{CFIC}~\citep{DBLP:journals/corr/abs-2402-09760}, \textbf{CRAG}~\citep{DBLP:journals/corr/abs-2401-15884}, and \textbf{Self-RAG}~\citep{DBLP:journals/corr/abs-2310-11511}; 3) \textbf{Vanilla RAG} (only retriever $\mathcal{R}$ and generator $\mathcal{G}$) based on various LLMs.
These LLMs range from small parameter-size (6b\textasciitilde8b) models
like ChatGLM3-6B-32k
\citep{DBLP:conf/acl/DuQLDQY022},
Qwen1.5-7b-32k
\citep{DBLP:journals/corr/abs-2309-16609}, 
Vicuna-v1.5-7b-16k
\citep{DBLP:conf/nips/ZhengC00WZL0LXZ23},
and Llama3-8B-8k
\citep{DBLP:journals/corr/abs-2302-13971}
to large parameter-size online models like GPT-3.5-Turbo\footnotemark[3] (\texttt{gpt-3.5-turbo-0125}) and GLM-4\footnotemark[4] (\texttt{glm-4}). 
\footnotetext[3]{\url{https://openai.com/blog/chatgpt}}
\footnotetext[4]{Due to resource limitations, we perform the API of glm4 with an 8k token window. \url{https://open.bigmodel.cn}.}

\noindent\textbf{Others.}
In our experiments, all token lengths are measured by ChatGLM tokenizer. We evaluate four different retrieval strategies to analyze the performance of LongRAG comprehensively (more details and results in Appendix \ref{Results of Different Retrieval Strategies}).
Specifically, we represent four retrieval strategies as "chunk size*top-$k$", including "200*7", "200*12", "500*3", and "500*5".
By default, we set the chunk size to 200 words and the top-$k$ value to 7.
\subsection{Overall Performance}
In this section, we perform a multi-dimensional comparison and analysis of the overall performance results in Table~\ref{tab:overall}.

\noindent\textbf{Ours vs. Long-Context LLM Methods.}
We align the parameter size of \textit{Llama2} and compare LongRAG with the results of LongAlign and LongLoRA. Our system paradigm using SFT achieves the highest performance on all datasets.
In addition, we also observe that the LongRAG system paradigm equiping other similar parameter-size LLMs consistently surpasses baselines within Long-context LLM methods across all datasets.
These achievements confirm the superiority of our system across all datasets.
This occurs because long-context LLMs often overlook crucial factual details in the middle, while LongRAG precisely and robustly perceives factual details.
Overall, our system serves as a more effective technical solution for LCQA.

\noindent\textbf{Ours vs. Other RAG.}
We compare LongRAG with two categories of RAG baselines, advanced RAG and Vanilla RAG (RAG-Base, R\&B).
We employ the LangGraph library\footnotemark[5], integrated within the LangChain framework, to reproduce Self-RAG and CRAG.
\footnotetext[5]{\url{https://github.com/langchain-ai/langgraph}}
\textbf{First}, compared to the advanced RAG, especially Self-RAG, our LongRAG achieves a 6.16\% improvement across three datasets on average.
This is due to the self-reflective chain decision-making in Self-RAG, which can, in certain cases, amplify decision errors, leading to catastrophic loss of factual details. Similarly, CRAG exhibits non-robust evaluation behaviors, making it challenging to handle complex, multi-hop long-context questions.
\textbf{Second}, compared to the R\&B, all LLMs applied in our system exhibit significant improvements (up to 17.25\%).
Vanilla RAG segments long contexts into smaller semantic units, hindering the downstream generator from accessing a more coherent long-context background and the original long-context structure.
Based on the above analysis, our system, after performing extractor and filter, acquires higher-quality and less noise knowledge, thus generating more accurate answers.

\noindent\textbf{Small-Size vs. Large-Size LLMs.}
We find that the LongRAG system paradigm, whether employing fine-tuned small-size or non-fine-tuned large-size LLMs, consistently outperforms other baseline methods across all datasets.
Most importantly, LongRAG using the fine-tuned ChatGLM3-6B-32k achieves better performance than using non-fine-tuned GPT-3.5-Turbo.
These results prove our system paradigm boosts the ability to analyze and process complex long contexts, as well as "\textit{instruction following}" capability. It also compensates for the limitations observed in small-size LLMs, particularly in long-context in-context learning (ICL) and understanding complex information.

\begin{table*}[t]
  \centering
    \setlength{\tabcolsep}{2.5pt}
    \fontsize{8}{9}\selectfont
    \begin{tabular}{cccccc|ccccc|ccccc}
    \toprule
    \multirow{2}{*}{Model} & \multicolumn{5}{c}{\textbf{HotpotQA}} & \multicolumn{5}{c}{\textbf{2WikiMQA}} & \multicolumn{5}{c}{\textbf{MusiQue}} \\
\cmidrule{2-16}      & R\&B & R\&L & Ext. & Fil. & E\&F & R\&B & R\&L & Ext. & Fil. & E\&F & R\&B & R\&L & Ext. & Fil. & E\&F \\
    \rowcolor{gray!20}
    \multicolumn{16}{c}{\# \textbf{Ours} with SFT \#} \\
    LongRAG-ChatGLM3-6B-32k & 51.48  & 54.00  & \underline{55.11}  & 49.01  & \textbf{55.93}  & 46.61  & 44.83  & \underline{52.53}  & 48.83  & \textbf{54.85}  & 24.02  & \textbf{33.15}  & 32.98  & 27.70  & \underline{33.00}  \\
    LongRAG-Qwen1.5-7B-32k & 47.09  & 48.93  & \underline{50.01}  & 49.11  & \textbf{52.91}  & 35.78  & 37.72  & \underline{42.91}  & 38.98  & \textbf{46.65}  & 20.68  & 26.08  & \underline{29.60}  & 23.67  & \textbf{31.85}  \\
    LongRAG-Vicuna-v1.5-7B-16k & 51.63  & 50.18  & \textbf{55.94}  & 52.34  & \underline{55.55}  & 39.45  & 43.53  & \underline{49.57}  & 41.18  & \textbf{50.13}  & 25.30  & 25.28  & \underline{29.25}  & \textbf{29.29}  & 28.29  \\
    LongRAG-Llama3-8B-8k & 49.45  & 50.49  & \underline{51.77}  & 49.64  & \textbf{52.39}  & 39.79  & 37.16  & \underline{46.80}  & 42.40  & \textbf{49.67}  & 21.41  & 22.90  & \textbf{33.85}  & 23.47  & \underline{31.70}  \\
    \rowcolor{gray!20}
    \multicolumn{16}{c}{\# \textbf{Ours} without SFT \#} \\
    LongRAG-ChatGLM3-6B-32k & \underline{52.57}  & 50.19  & 52.27  & \textbf{53.36}  & 52.07  & 42.56  & 42.92  & \underline{44.95}  & 42.94  & \textbf{46.08}  & 25.51  & \textbf{29.93}  & 28.27  & 23.99  & \underline{28.45}  \\
    LongRAG-Qwen1.5-7B-32k & 45.70  & 49.72  & \underline{50.74}  & 45.70  & \textbf{50.80}  & 34.69  & \underline{35.49}  & \textbf{39.53}  & 34.69  & \textbf{39.53}  & \underline{25.08}  & 25.85  & \textbf{29.75}  & \underline{25.08}  & \textbf{29.75}  \\
    LongRAG-Vicuna-v1.5-7B-16k & 38.63  & 30.40  & \underline{41.45}  & 39.46  & \textbf{43.18}  & 27.92  & 20.68  & \underline{29.08}  & 29.89  & \textbf{30.85}  & 15.68  & 8.92  & \textbf{17.65}  & \underline{16.35}  & 16.98  \\
    LongRAG-Llama3-8B-8k & 48.25  & 48.72  & \textbf{52.44}  & 47.75  & \underline{52.19}  & 43.47  & 41.59  & \textbf{47.34}  & 42.22  & \underline{46.57}  & 19.66  & 23.62  & \underline{24.90}  & 20.06  & \textbf{24.99}  \\
    LongRAG-GPT-3.5-Turbo & 52.31  & 55.30  & \underline{56.15}  & 50.90  & \textbf{56.17}  & 43.44  & 45.03  & \textbf{53.29}  & 39.49  & \underline{51.37}  & 25.22  & 28.65  & \underline{32.17}  & 24.41  & \textbf{32.83}  \\
    LongRAG-GPT-3.5-Turbo-16k & 50.17  & 49.80  & \textbf{60.06}  & 47.10  & \underline{59.11}  & 45.32  & 46.80  & \textbf{51.26}  & 46.38  & \underline{51.25}  & 21.84  & 25.09  & \underline{26.92}  & 22.02  & \textbf{30.37}  \\
    LongRAG-GLM-4 & 57.41  & 56.17  & \underline{61.07}  & 55.41  & \textbf{62.11}  & 52.91  & 48.98  & \underline{54.22}  & 52.61  & \textbf{57.16}  & 27.55  & 27.85  & \textbf{38.54}  & 28.12  & \underline{38.40}  \\
    \bottomrule
    \end{tabular}%
    \caption{\textbf{Results (\%) of the ablation study.} We compare five strategies in two dimensions: with and without SFT.
    We highlight the highest ("\textbf{Bold Font}") and second-highest ("\textbf{\_}") results per model. R\&B, R\&L, Ext., Fil., and E\&F represent RAG-Base, RAG-Long, Extractor, Filter, and Extractor \& Filter, respectively.}
  \label{tab:ablation}%
\end{table*}%
\subsection{Ablation Study}
The ablation study (Table~\ref{tab:ablation}) reports results within five strategies to highlight the effectiveness of the information extractor and CoT-guided filter. In the following paragraphs, we explore the reasons for the performance gains.

\noindent\textbf{RAG-Long vs. RAG-Base.}
RAG-Long (\textbf{R\&L}) refers to mapping the $p_c$ back to the $p$ and then directly putting a set of $p$ into the generator to output a response.
The R\&L strategy fails to robustly achieve performance improvements over R\&B.
Specifically, the R\&L strategy feeds the continuous long-context space into the LLM, unlike the R\&B disrupts the semantic continuity of long contexts. Therefore, R\&L enables to capture of a broader continuity of the source semantic space; however, it also risks introducing excessive noise.

\noindent\textbf{Extractor vs. RAG-Long.}
The extractor builds upon the R\&L to effectively extract pertinent long-context information.
Specifically, the extractor strategy refers to the system first extracting global information $I_{g}$
from the mapped source long paragraphs, and then using $I_{g}$ as supplementary input alongside retrieved chunks $p_{c}$ to the generator to enhance answer quality.
The system using the extractor strategy presents substantial improvements across all three datasets, particularly on larger-size LLMs that exhibit stronger in-context learning capability.
This improvement stems from recognizing the challenge of directly deriving answers from lengthy contexts; therefore, we first leverage the LLMs' capability to extract global structures and background knowledge as supplements for generating the final answer.
The extractor strategy effectively mitigates the issue of low-quality responses in the R\&L strategy caused by directly feeding redundant long passages into LLMs, while also providing LLMs with additional and concise global structure and contextual relationship information.
Additionally, in most instances, the extractor is the primary contributor to performance gains, second only to the joint strategy, Extractor \& Filter (\textbf{E\&F}).

\noindent\textbf{Filter vs. RAG-Base.}
Using the filter alone based on R\&B improves the performance only marginally in a few cases.
This occurs because filtering is, after all, a process of information reduction.
Therefore, it can only display markedly performance when used in conjunction with the Extractor.

\noindent\textbf{Extractor \& Filter vs. Others.}
E\&F serves as a joint strategy with two pluggable components within the RAG system, achieving the best performance in the majority of cases.
It outperforms the R\&L strategy by providing refined information with less noise, thereby effectively alleviating the "\textit{lost in the middle}" issue.
Specifically, the role of the Extractor is to capture globally effective information from long contexts, while the Filter flexibly selects factual details through interactions between the question and relevant paragraphs.
Results suggest employing both E\&F components yields a more helpful and concise set of information compared to using a single component.
However, it is worth mentioning that a minority of cases where E\&F underperforms compared to Extractor alone do not imply that the Filter is ineffective.
In fact, when the built-in LLM possesses strong "\textit{instruction-following}" capabilities (e.g., GLM-4 and fine-tuned small-size LLMs), adding the Filter is more likely to boost system performance. Plus, the Filter can reduce the number of tokens input into downstream LLMs. From the results in Table~\ref{tab:ablation} and Figure~\ref{fig3:token analysis}, it is evident that using the Filter can save token costs during the generation phase while achieving performance comparable to or even better than using the Extractor alone.
Furthermore, we find that not all researchers can afford the high costs of powerful API LLMs (e.g., GPT-3.5-Turbo). Our method offers an alternative by using more affordable open-source local LLMs for components before the generator, instead of relying on expensive online APIs throughout the entire inference process. Therefore, if the goal is to balance performance and cost, E\&F is crucial.

\subsection{Discussion}
\begin{figure}[t]
  \includegraphics[width=\columnwidth]{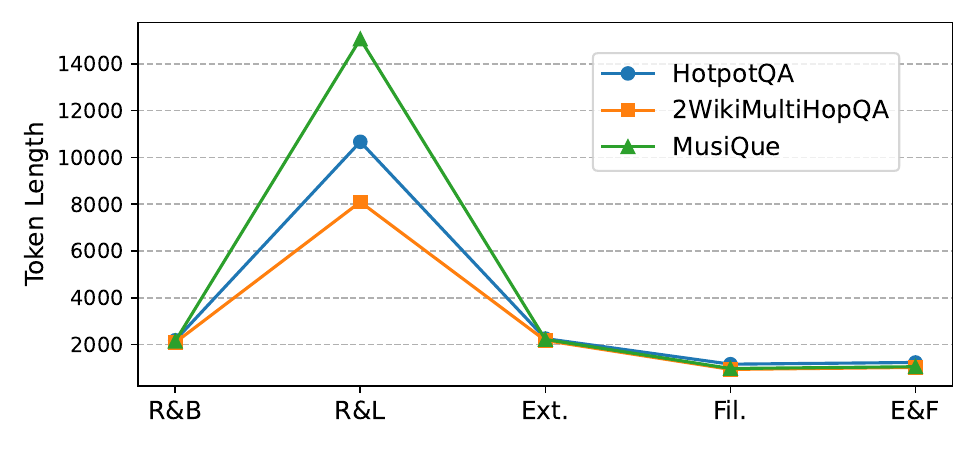}
  \caption{Trends of token lengths fed into the Generator $\mathcal{G}$ of five component strategies on three datasets.}
  \label{fig3:token analysis}
\end{figure}

\noindent\textbf{Analysis of Token Length Trends.}
Figure~\ref{fig3:token analysis} illustrates the token lengths inputted into the generator $\mathcal{G}$
for all datasets
after undergoing the five strategies.
The results indicate a consistent trend across all datasets.
Specifically, our E\&F strategy feeds $\mathcal{G}$ fewer tokens but achieves superior outcomes, however, R\&L feeds the most without corresponding systematic gains, which indicates we can obtain higher quality information through E\&F.

\begin{figure}[t]
  \includegraphics[width=\columnwidth]{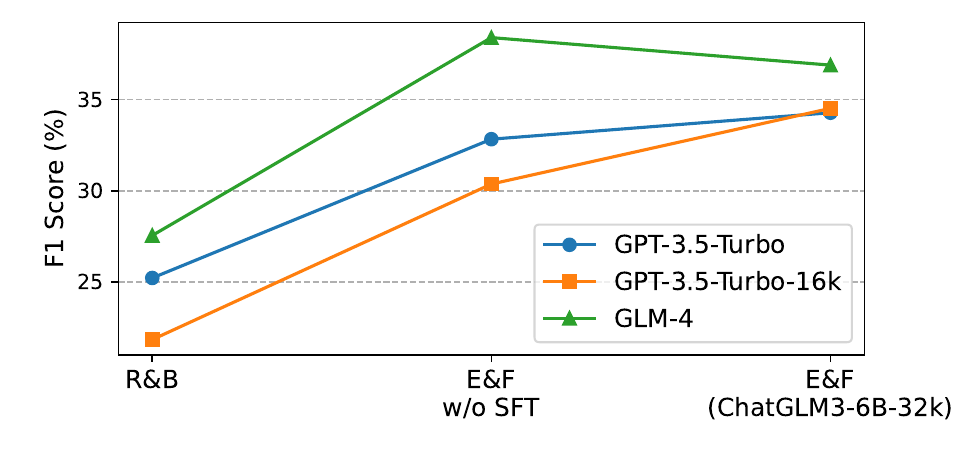}
  \caption{Analysis of the transferability of Extractor\&Filter on dataset MusiQue.}
  \label{fig4:Analysis of the transferability of E&F.}
\end{figure}

\noindent\textbf{Component Transferability.}
\label{discussion-2}
As shown in Figure \ref{fig4:Analysis of the transferability of E&F.}, 
E\&F (ChatGLM3-6B-32k) means we employ ChatGLM3-6B-32k as the built-in LLM of extractor $\mathcal{E}$ and filter $\mathcal{F}$, while the generator $\mathcal{G}$ uses other powerful online LLMs, e.g., GPT-3.5-Turbo.
E\&F w/o SFT represents the same meanings in Table~\ref{tab:ablation}, that is, we apply the same built-in LLM for the $\mathcal{E}$, $\mathcal{F}$, and $\mathcal{G}$. Results reveal we transfer the expensive powerful online LLMs of $\mathcal{E}$ and $\mathcal{F}$ to a low-cost local model while achieving excellent results. It can surpass GPT-3.5-Turbo and rival the GLM-4.

\section{Conclusion}
We build an effective and robust RAG system paradigm — \textbf{LongRAG} — which enhances RAG's performance in LCQA tasks via a dual information perspective. LongRAG addresses two main issues faced by existing methods: 1) the incomplete collection of long-context information; and 2) the difficulty in precisely identifying factual information amid substantial noise. 
We conduct extensive multi-dimensional experiments, which demonstrate the superiority of LongRAG and the effectiveness of our proposed components and fine-tuning strategy. 
LongRAG significantly outperforms long-context LLMs, advanced RAG methods, and Vanilla RAG based on various LLMs.
Our plug-and-play components successfully use small parameter-size LLMs, replacing expensive online API resources with low-cost local deployment solutions, while better than GPT-3.5-Turbo.
Additionally, we provide an automated pipeline for fine-tuning instruction data construction, which greatly facilitates the application of our system to other specific-domain data.

\section{Limitations}
This paper presents a general-purpose and corpus-level retrieval-augmented generation system paradigm for long-context question answering, termed LongRAG. While the system paradigm brings significant advancements and proves effective, it is also subject to certain limitations that merit discussion.

\noindent\textbf{One-time Retrieval Dependency.}
In this study, we only investigated the performance of the information extractor and CoT-guided filter in a one-time retrieval scenario.
The quality of CoTs and source documents for answering depends on the quality of single-pass retrieved chunks.
Consequently, low-quality one-time retrieval can indirectly undermine the effectiveness of our core components.
Moving forward, we anticipate that an effective avenue of improvement could develop an adaptive multi-round retrieval strategy through interaction with core components.

\noindent\textbf{Dataset Annotation Bias.}
Although we have used the 32-billion parameter ChatGLM3 model to generate high-quality fine-tuning datasets, models of this scale may still be susceptible to annotation biases inherent in self-generated datasets.
Such biases could impair the contextual understanding of the fine-tuned models across diverse tasks and domains, potentially undermining the overall system performance.
It is therefore valuable to thoroughly investigate the performance of instruction datasets created by LLMs of various scales in cross-domain and multi-task environments.
\section*{Acknowledgments}
This work is supported by the Natural Science Foundation of China (NSFC) 62276148 and 62425601, Tsinghua University (Department of Computer Science and Technology) -Siemens Ltd., China Joint Research Center for Industrial Intelligence and Internet of Things (JCIIOT) and New Cornerstone Science Foundation through the XPLORER PRIZE.

\bibliography{acl}

\clearpage
\appendix
\section{Additional Experimental Results}
\label{Additional Experimental Results}

\subsection{Results of Different Retrieval Strategies}
\label{Results of Different Retrieval Strategies}
Table~\ref{tab:Overall HotpotQA}, Table~\ref{tab:Overall 2WikiMultiHopQA}, and Table~\ref{tab:Overall MusiQue} display all of overall performance results. We evaluate four different retrieval strategies to analyze the performance of LongRAG comprehensively.
These strategies include 200*7, 200*12, 500*3, and 500*5.
For example, "200*7" stands for "chunk size*top-$k$".
By comparing these retrieval strategies, we observe that an intermediate value for the top-k setting tends to yield superior performance. This phenomenon arises from the extractor's utilization of the source long paragraphs mapped from top-$k$ recalled chunks. Too few recalled chunks may result in insufficient collection of extensive contextual information, while an excessive number may introduce more noise. Contrasting the outcomes of 200*7 and 500*3, we notice that, under comparable context length,
a smaller chunk size coupled with a higher top-k recall number can maximize the acquisition of global information within the corpus space, thereby exhibiting enhanced performance. These results confirm the efficacy of the core components ($\mathcal{E}$ and $\mathcal{F}$) in our system.

\subsection{Component Transferability}
We provide specific values in Figure~\ref{fig4:Analysis of the transferability of E&F.} in section~\ref{discussion-2} with experimental results (Table~\ref{tab:modelmove:HotpotQA}, Table~\ref{tab:modelmove:2WikiMultiHopQA} and Table~\ref{tab:modelmove:MusiQue}) for
all datasets, including HotpotQA, 2WikiMultiHopQA, and MusiQue.
In 2WikiMultiHopQA and HotpotQA, our system also exhibits component transferability similar to that in MusiQue. We conducted all experiments using ChatGLM3-6B-32k with SFT as a relatively low-cost local model.

\subsection{Analysis of Token Length Trends}
Figure~\ref{fig3:token analysis} only shows the token length trend using ChatGLM3-6B-32k with SFT across five strategies. The specific values and the results of using more built-in fine-tuned LLMs are shown in Table~\ref{tab:token_HotpotQA}, Table~\ref{tab:token_2WikiMultiHopQA}, and Table~\ref{tab:token_MusiQue}.

\subsection{Additional Baseline Results}
As an agent framework, ReAct can also be instantiated as an efficient RAG system based on adaptive retrieval~\citep{DBLP:conf/iclr/YaoZYDSN023}. ReAct can answer questions through the process of "\textit{Thought/Action/Observation}".
In our experiment, we define "\textit{Action}" as the retrieval action, meaning that when knowledge needs to be retrieved, the relevant information is retrieved from our local corpus $\mathcal{C}$.
We have aligned the experimental parameters, and the results of the ReAct experiment are presented in Table~\ref{tab:Results of ReAct}.
\begin{table}[t]
  \centering
    \setlength{\tabcolsep}{8pt}
    \fontsize{9}{11}\selectfont
    \begin{tabular}{cc}
    \toprule
    \textbf{Datasets} & ReAct (\textit{GPT-3.5-Turbo}) \\   
    \midrule
    HotpotQA & 49.60 \\
    2WikiMultihopQA & 41.86 \\
    MuSiQue & 27.81 \\
    \midrule
    Average & 39.76 \\
    \bottomrule
    \end{tabular}%
    \caption{Results of ReAct.}
  \label{tab:Results of ReAct}%
\end{table}%

\section{Experimental Details Explanation}
\subsection{Details of Baseline Replication}
Self-RAG, CRAG, LongLoRA, and LongAlign produce too long responses, making it challenging to fairly compare them with our method using the F1-score as an evaluation metric. In other words, the long outputs result in lower scores for these baselines. Therefore, we select the LLM with a strong ability of "\textit{instruction-following}", such as GPT-3.5-Turbo, and perform few-shot ICL on their outputs to produce the final answers.
In the following paragraphs, we will introduce the specific experimental details involved in reproducing the results for Self-RAG and CRAG.

We employ the LangGraph library, integrated within the LangChain framework, to reproduce Self-RAG and CRAG.
Specifically, Self-RAG employs an adaptive retrieval based on self-reflection. If the LLM identifies the retrieved chunks as irrelevant, or the generated outputs are regarded as unanswerable, Self-RAG will restart the search and answer process until the maximum number of rounds. 
In our experiments, we set the maximum number of retrieval rounds to 3.
If, upon reaching this round limit,
all retrieved documents are still considered irrelevant, there are two answer strategies:
The first strategy uses all chunks retrieved during the final round, while the second strategy involves answering without using the retrieved chunks.
In Table~\ref{tab:overall} of the main paper, we report the results of the first strategy, which shows higher results than those of the second strategy. Additionally, we present the performance of the second strategy in Table~\ref{tab:Results of Self-RAG via the second strategy.}.

\begin{table}[htbp]
  \centering
    \setlength{\tabcolsep}{8pt}
    \fontsize{9}{11}\selectfont
    \begin{tabular}{cc}
    \toprule
    \textbf{Datasets} & Self-RAG (\textit{GPT-3.5-Turbo}) \\   
    \midrule
    HotpotQA & 44.99 \\
    2WikiMultihopQA & 19.79 \\
    MuSiQue & 23.49 \\
    \bottomrule
    \end{tabular}%
    \caption{Results of Self-RAG via the second strategy.}
  \label{tab:Results of Self-RAG via the second strategy.}%
\end{table}%

CRAG has implemented a fallback strategy to prevent a steep decline in response quality due to all retrieved chunks being filtered out. When the retrieved chunks are considered insufficient to answer the question, it is supplemented with external knowledge retrieved from the web. For a fair reproduction in our experiments, when faced with similar issues, we rewrite the question and conduct another retrieval from our corpus $\mathcal{C}$. Since our corpus contains all the relevant information necessary to answer the question, we do not need to retrieve external knowledge from the web.

\subsection{Details of the Corpus}
\label{Details of Experimental Datasets}
Our experimental datasets and the corpus used for knowledge retrieval are constructed based on LongBench. The multi-hop QA datasets of LongBench include questions, answers, and multiple corresponding paragraphs concatenated to form long contexts of each question. To adapt it for the RAG system, we split long contexts into individual corresponding paragraphs. Since each paragraph is a semantically coherent and complete Wikipedia paragraph, we treat each paragraph $p$ as an independent knowledge unit. After deduplication, the paragraphs from all questions form the corpus $\mathcal{C}$.

\section{Details of LRGinstruction}
\label{Details of LRGinstruction}
We construct an instruction dataset for fine-tuning, comprising four types of data, each designed to enhance the "\textit{instruction-following}" capability of corresponding components.
The four types of data include \textbf{Long-Context Extractor data}, \textbf{CoT-guiding Data}, \textbf{Filtering Data}, and \textbf{Task-Oriented Data}.
To be specific, long-context extractor data is utilized to enhance the capabilities of the LLM-augmented extractor. CoT-guiding data and filtering data are applied to strengthen the abilities of the two-stage CoT-guided filter.
Question and answer data are utilized to enhance the generator's capability, learning the specific answering style required for tasks.
We present examples of all the pipelines used for data construction and formats of the generated data (golden data) in Table~\ref{Data construction prompt for Extractor}, Table~\ref{Data construction prompt for Filter} and Table~\ref{Data construction prompt for RAG task}.
Specific examples of four types of golden data are also shown in Table~\ref{tab:construction dataset: cite}, Table~\ref{tab:construction dataset: CoT}, Table~\ref{tab:construction dataset: Filter} and Table~\ref{tab:construction dataset: RAG task}.
To clearly distinguish between prompts for data construction and generated instruction data, we mark prompts in each pipeline as \textbf{[STEP]} and instruction data as \textbf{[RESULT]}. 
The following paragraphs will elaborate on the construction details and pipelines.

\subsection{Data Pre-Processing}
\label{Appendix: Data Pre-Processing}
We further detail the random strategy.
The number of distracting paragraphs $P_{d}$ in our instruction data is randomly chosen within a specific range, from two up to the total length of 
$P_{d}$, mathematically expressed as $[2, maxLen(P_{d})]$.
Moreover, we further detail how to discard any question-answer pairs with insufficient context length.
Here, "\textit{insufficient context length}" means that the total token length of all corresponding paragraphs provided for a question is lower than a specific threshold. Specifically, we use a threshold of 1.5k for HotpotQA and 2WikiMultiHopQA, and 2.5k for MusiQue. During the experiment, we find that this threshold setting preserves long-context samples, enabling the model to learn long-context styles and retain sufficient data for training. For QASPER, we do not filter any samples because the papers are inherently long.

\subsection{Long-Context Extractor Data}
\label{Appendix: Long-Context Extractor Data}
In the construction pipeline (Table~\ref{Data construction prompt for Extractor}) for LLM-augmented extractor data, we aim to feed the question and $P_{s}$ into the LLM, which outputs all the relevant information for answering the question.
We provide the specific construction process and details shown in Table~\ref{Data construction prompt for Extractor}.
We construct the initial dataset via \textbf{[STEP-1]}, which {global information} as gold outputs.
If the response of \textbf{[STEP-1]} is particularly short, we discard it due to a small amount of effective information, with a discard threshold of 20 tokens.
Subsequently, in \textbf{[STEP-2]}, we perform a self-evaluator of the gold output after \textbf{[STEP-1]}. Only samples that pass the validation (i.e., those for which the output in \textbf{[STEP-2]} is "\texttt{True}") are included in the final instruction dataset. The final \textbf{[RESULT]} presents the ultimate gold data (long-context extractor data) in this pipeline, and "\texttt{\{\textcolor{blue}{content}\}}" represents $P$ including both $P_{s}$ and selected $P_{d}$ by random strategy.
This type of data enhances the LLM-augmented extractor to identify valuable evidence information from substantial lengthy context source paragraphs.

\subsection{CoT-guiding Data \& Filtering Data}
\label{Appendix: CoT-guiding Data and Filtering Data}
In the CoT-guided filter, we employ a two-stage strategy to precisely and flexibly screen problem-related chunks while discarding redundant chunks. 
The two types of data, CoT-guiding data  (\textbf{[RESULT-1]})  and filtering data (\textbf{[RESULT-2]})  aim to enhance the "\textit{instruction-following}" ability of the two-stage components of the CoT-guided filter, and better identify factual details.
This construction pipeline and final constructed data are shown in Table~\ref{Data construction prompt for Filter}. 
First, in \textbf{[STEP-1]}, we generate a guiding CoT by inputting the question and all corresponding $P_{s}$. The generated CoT provides global clues for question-answering by performing in-context learning in all retrieved chunks.
If the CoT is particularly short, we consider it a low-quality clue and discard it, with a discard threshold of 20 tokens.
In \textbf{[STEP-2]}, we then perform a self-evaluator of the guiding CoT \textbf{[STEP-1]} to verify the feasibility of the CoT in responding to the question. In the self-evaluator, we use the answers from the raw dataset as the basis for judging the quality of CoT.
\textbf{[RESULT-1]} displays the instruction data constructed for the CoT-guided stage, named CoT-guiding data, and "\texttt{\{\textcolor{blue}{content}\}}" represents $P$ including both $P_{s}$ and selected $P_{d}$ by random strategy.
Finally, for the filtering stage, we treat each paragraph $p$ as a unit and regard given binary discrete labels in the raw dataset as gold labels, expressed as \texttt{\{\textcolor{blue}{status}\}} in \textbf{[RESULT-2]}. 
The filtering stage instruction data is shown in \textbf{[RESULT-2]}. Its "\texttt{\{\textcolor{blue}{content}\}}" represents each paragraph $p\in P$.
It is worth noting that in the original dataset, the number of $p$ marked as "\texttt{True}" is much lower than "\texttt{False}". To ensure the uniformity of the distribution, we select 100 samples with a status of "\texttt{True}" and 100 samples with a status of "\texttt{False}".

\subsection{Task-Oriented Data}
\label{Appendix: Task-Oriented Data}
The questions and answers are already provided in the original datasets. We standardize their format to construct the question-answering data (see Table~\ref{Data construction prompt for RAG task}) in our fine-tuning instruction dataset. The "\texttt{\{\textcolor{blue}{content}\}}" in \textbf{[RESULT]} represents $P$ including both $P_{s}$ and selected $P_{d}$ by random strategy.
\subsection{Statistics of  LRGinstruction}
\label{Appendix: Statistics of  LRGinstruction}
To sum up, we derive four types of data from the training sets of the HotpotQA, 2WikiMultiHopQA, and MusiQue datasets, with each type of data containing 200 samples. This results in 800 samples per dataset and a total of 2400 samples across the three datasets. The token length of each instruction data is less than 7k.
Furthermore, to adapt our RAG system to long-context QA, we also derive two types of data (i.e., long-context extractor data and CoT-guiding data) using the QASPER dataset, each type of data with 100 samples, and each instruction data length ranging from 6k-29k.
We list the statistics of our fine-tuning instruction dataset in Table~\ref{statistics of our fine-tuning instruction dataset}.
\section{Prompts of LongRAG System}
\label{Prompts of LongRAG system}
We present all prompts in LongRAG's components in Table~\ref{tab:Prompts of LongRAG system}.
The "\texttt{\{\textcolor{blue}{content}\}}" in different prompts represent different contextual information. To be specific, the "\texttt{\{\textcolor{blue}{content}\}}" in the prompt of LLM-augmented information extraction represents all source long-context paragraphs $p$ after the mapping strategy. In the prompt of the CoT guidance stage in the CoT-guided filter, it represents all retrieval chunks $p_c$, while in the prompt of the filtering stage, it represents each $p_c$.

\section{Answer Examples}
\label{LongRAG Examples}
We provide answer examples shown in Table~\ref{tab:System Case: RAG-base}, Table~\ref{tab:System Case: RAG-Long}, and Table~\ref{tab:System Case: Extractor and Filter}. LongRAG addresses the issues of incomplete information and "\textit{lost in the middle}" found in Vanilla RAG and RAG-Long, while requiring fewer tokens inputted into the generator yet showing superior response performance.

\clearpage
\begin{table*}[h]
  \centering
    \begin{tabular}{ccccc}
    \toprule
    Model & \multicolumn{4}{c}{\textbf{HotpotQA}} \\
    \cmidrule{2-5}
    & 200*7 & 200*12 & 500*3 & 500*5 \\
    \midrule
    \multicolumn{5}{c}{\#  RAG Base (Vanilla RAG)  \#} \\
    \midrule
    ChatGLM3-6B-32k & 52.57 & 53.10 & 47.72 & 51.17 \\
    Qwen1.5-7B-32k & 45.70 & 49.20 & 44.43 & 44.16 \\
    Vicuna-v1.5-7B-16k & 38.63 & 34.35 & 37.23 & 35.32 \\
    Llama3-8B-8k & 48.25 & 51.69 & 47.12 & 50.88 \\
    GPT-3.5-Turbo & 52.31 & 55.21 & 52.84 & 51.21 \\
    GPT-3.5-Turbo-16k & 50.17 & 53.58 & 48.02 & 48.84 \\
    Llama3-70B-8k & 52.33 & 53.53 & 49.51 & 51.38 \\
    GLM-4 & 57.41 & 59.55 & 53.71 & 58.45 \\
    \midrule
    \multicolumn{5}{c}{\# \textbf{Ours} with SFT \#} \\
    \midrule
    LongRAG-ChatGLM3-6B-32k & 55.93 & 54.36 & 50.72 & 54.67 \\
    LongRAG-Qwen1.5-7B-32k & 52.91 & 52.27 & 49.70 & 50.69 \\
    LongRAG-Vicuna-v1.5-7B-16k & 55.55 & 54.79 & 52.26 & 52.89 \\
    LongRAG-Llama3-8B-8k & 52.39 & 52.00 & 49.05 & 54.62 \\
    \midrule
    \multicolumn{5}{c}{\# \textbf{Ours} without SFT \#} \\
    \midrule
    LongRAG-GPT-3.5-Turbo & 56.17 & 56.06 & 55.63 & 55.11 \\
    LongRAG-GPT-3.5-Turbo-16k & 59.11 & 51.55 & 48.45 & 55.57 \\
    LongRAG-GLM-4 & 62.11 & 60.55 & 55.36 & 61.14 \\
    \bottomrule
    \end{tabular}%
    \caption{Overall performance of our LongRAG on HotpotQA dataset.}
  \label{tab:Overall HotpotQA}%
\end{table*}%

\clearpage
\begin{table*}[h]
  \centering
    \begin{tabular}{ccccc}
    \toprule
    Model & \multicolumn{4}{c}{\textbf{2WikiMultiHopQA}} \\
    \cmidrule{2-5}      & 200*7 & \multicolumn{1}{l}{200*12} & \multicolumn{1}{l}{500*3} & \multicolumn{1}{l}{500*5} \\
    \midrule
    \multicolumn{5}{c}{\#  RAG Base (Vanilla RAG)  \#} \\
    \midrule
    ChatGLM3-6B-32k & 42.56 & 38.71 & 40.65 & 42.34 \\
    Qwen1.5-7B-32k & 34.69 & 34.79 & 34.47 & 35.24 \\
    Vicuna-v1.5-7B-16k & 27.92 & 26.39 & 32.76 & 26.36 \\
    Llama3-8B-8k & 43.47 & 40.01 & 30.48 & 41.44 \\
    GPT-3.5-Turbo & 43.44 & 40.06 & 43.17 & 39.69 \\
    GPT-3.5-Turbo-16k & 45.32 & 39.09 & 43.31 & 42.49 \\
    Llama3-70B-8k & 50.23 & 48.91 & 46.61 & 50.10 \\
    GLM-4 & 52.91 & 52.37 & 49.48 & 51.06 \\
    \midrule
    \multicolumn{5}{c}{\# \textbf{Ours} with SFT \#} \\
    \midrule
    LongRAG-ChatGLM3-6B-32k & 54.85 & 58.51 & 49.28 & 53.51 \\
    LongRAG-Qwen1.5-7B-32k & 46.65 & 45.23 & 42.96 & 44.55 \\
    LongRAG-Vicuna-v1.5-7B-16k & 50.13 & 50.93 & 47.45 & 48.02 \\
    LongRAG-Llama3-8B-8k & 49.67 & 51.41 & 43.80 & 49.70 \\
    \midrule
    \multicolumn{5}{c}{\# \textbf{Ours} without SFT \#} \\
    \midrule
    LongRAG-GPT-3.5-Turbo & 51.37 & 56.55 & 48.16 & 48.60 \\
    LongRAG-GPT-3.5-Turbo-16k & 51.25 & 45.45 & 44.08 & 44.21 \\
    LongRAG-GLM-4 & 57.16 & 52.90 & 44.93 & 50.05 \\
    \bottomrule
    \end{tabular}%
    \caption{Overall performance of our LongRAG on 2WikiMultiHopQA dataset.}
  \label{tab:Overall 2WikiMultiHopQA}%
\end{table*}%

\clearpage
\begin{table*}[h]
  \centering
    \begin{tabular}{ccccc}
    \toprule
    Model & \multicolumn{4}{c}{\textbf{MusiQue}} \\
\cmidrule{2-5}      & 200*7 & 200*12 & 500*3 & 500*5 \\
\midrule   \multicolumn{5}{c}{\#  RAG Base (Vanilla RAG)  \#} \\
    \midrule
    ChatGLM3-6B-32k & 25.51  & 25.91  & 24.31  & 25.63  \\
    Qwen1.5-7B-32k & 25.08  & 23.51  & 21.08  & 22.05  \\
    Vicuna-v1.5-7B-16k & 15.68  & 14.55  & 16.05  & 13.89  \\
    Llama3-8B-8k & 19.66  & 23.65  & 19.33  & 22.51  \\
    GPT-3.5-Turbo & 25.22  & 28.23  & 25.34  & 27.06  \\
    GPT-3.5-Turbo-16k & 21.84  & 25.41  & 24.80  & 23.79  \\
    Llama3-70B-8k & 25.49 & 27.72 & 23.05 & 24.13 \\
    GLM-4 & 27.55  & 33.93  & 27.92  & 27.56  \\
    \midrule
    \multicolumn{5}{c}{\# \textbf{Ours} with SFT \#} \\
    \midrule
    LongRAG-ChatGLM3-6B-32k & 33.00  & 33.12  & 30.09  & 31.98  \\
    LongRAG-Qwen1.5-7B-32k & 31.85  & 32.22  & 27.25  & 25.84  \\
    LongRAG-Vicuna-v1.5-7B-16k & 28.29  & 33.76  & 29.42  & 29.89  \\
    LongRAG-Llama3-8B-8k & 31.70  & 38.19  & 33.90  & 29.57  \\
    \midrule
    \multicolumn{5}{c}{\# \textbf{Ours} without SFT \#} \\
    \midrule
    LongRAG-GPT-3.5-Turbo & 32.83  & 32.64  & 29.83  & 28.03  \\
    LongRAG-GPT-3.5-Turbo-16k & 30.37  & 32.11  & 28.96  & 26.58  \\
    LongRAG-GLM-4 & 38.40  & 39.68  & 34.67  & 33.05  \\
    \bottomrule
    \end{tabular}%
    \caption{Overall performance of our LongRAG on MusiQue dataset.}
  \label{tab:Overall MusiQue}%
\end{table*}%

\clearpage
\begin{table*}[h]
  \centering
    \begin{tabular}{cccc}
    \toprule
    \multicolumn{1}{c}{\multirow{2}[4]{*}{\textbf{
    Generator}}} & \multicolumn{3}{c}{\textbf{HotpotQA}} \\
\cmidrule{2-4}      & R\&B & \multicolumn{1}{p{10em}}{\centering E\&F w/o SFT} & \multicolumn{1}{p{9em}}{\centering \makecell{E\&F w/ SFT \\ (ChatGLM3-6b-32k)}} \\
    \midrule
    LongRAG-GPT-3.5-Turbo-16k & 50.17  & 59.11  & 57.82  \\
    LongRAG-GPT-3.5-Turbo & 52.31  & 56.17  & 59.09  \\
    LongRAG-GLM-4 & 57.41  & 62.11  & 59.20 \\
    \bottomrule
    \end{tabular}%
    \caption{Analysis of the component transferability of E\&F on HotpotQA dataset.}
  \label{tab:modelmove:HotpotQA}%
\end{table*}%

\begin{table*}[h]
  \centering
    \begin{tabular}{cccc}
    \toprule
    \multicolumn{1}{c}{\multirow{2}[4]{*}{\textbf{
    Generator}}} & \multicolumn{3}{c}{\textbf{2WikiMultiHopQA}} \\
\cmidrule{2-4}      & R\&B & \multicolumn{1}{p{10em}}{\centering E\&F w/o SFT} & \multicolumn{1}{p{9em}}{\centering \makecell{E\&F w/ SFT \\ (ChatGLM3-6b-32k)}} \\
    \midrule
    LongRAG-GPT-3.5-Turbo-16k & 45.32  & 51.25  & 57.86  \\
    LongRAG-GPT-3.5-Turbo & 43.44  & 51.37  & 54.62  \\
    LongRAG-GLM-4 & 52.91  & 57.16  & 55.96 \\
    \bottomrule
    \end{tabular}%
    \caption{Analysis of the component transferability of E\&F on 2WikiMultiHopQA dataset.}
  \label{tab:modelmove:2WikiMultiHopQA}%
\end{table*}%

\begin{table*}[htbp]
  \centering
    \begin{tabular}{cccc}
    \toprule
    \multicolumn{1}{c}{\multirow{2}[4]{*}{\textbf{
    Generator}}} & \multicolumn{3}{c}{\textbf{MusiQue}} \\
\cmidrule{2-4}      & R\&B & \multicolumn{1}{p{10em}}{\centering E\&F w/o SFT} & \multicolumn{1}{p{9em}}{\centering \makecell{E\&F w/ SFT \\ (ChatGLM3-6b-32k)}} \\
    \midrule
    LongRAG-GPT-3.5-Turbo-16k & 21.84 & 30.37 & 34.52 \\
    LongRAG-GPT-3.5-Turbo & 25.22 & 32.83 & 34.28 \\
    LongRAG-GLM-4 & 27.55 & 38.40 & 36.89 \\
    \bottomrule
    \end{tabular}%
    \caption{Analysis of the component transferability of E\&F on MusiQue dataset.}
  \label{tab:modelmove:MusiQue}%
\end{table*}%

\clearpage
\begin{table*}[h]
  \centering
    \begin{tabular}{cccccc}
    \toprule
    \multirow{2}[4]{*}{Model} & \multicolumn{5}{c}{\textbf{HotpotQA}} \\
\cmidrule{2-6}      & R\&B & R\&L & Ext. & Fil. & E\&F \\
    \midrule
    LongRAG-ChatGLM3-6B-32k w/ SFT & 2181  & 10669  & 2254  & 1160  & 1233  \\
    LongRAG-Qwen1.5-7B-32k w/ SFT & 2181  & 10669  & 2248  & 1260  & 1327  \\
    LongRAG-Vicuna-v1.5-7B-16k w/ SFT & 2181  & 10596  & 2270  & 1233  & 1321  \\
    LongRAG-Llama3-8B-8k w/ SFT & 2181  & 7428  & 2243  & 1101  & 1163  \\
    \bottomrule
    \end{tabular}%
    \caption{Values of the token length fed into the generator on HotpotQA dataset.}
  \label{tab:token_HotpotQA}%
\end{table*}%

\begin{table*}[t]
  \centering
    \begin{tabular}{cccccc}
    \toprule
    \multirow{2}[4]{*}{Model} & \multicolumn{5}{c}{\textbf{2WikiMultiHopQA}} \\
\cmidrule{2-6}      & R\&B & R\&L & Ext. & Fil. & E\&F \\
    \midrule
    LongRAG-ChatGLM3-6B-32k w/ SFT & 2086  & 8096  & 2171  & 937  & 1022 \\
    LongRAG-Qwen1.5-7B-32k w/ SFT & 2086  & 8096  & 2162  & 941  & 1016  \\
    LongRAG-Vicuna-v1.5-7B-16k w/ SFT & 2086  & 8096  & 2176  & 937  & 1027  \\
    LongRAG-Llama3-8B-8k w/ SFT & 2086  & 6744  & 2150  & 813  & 876  \\
    \bottomrule
    \end{tabular}%
    \caption{Values of the token length fed into the generator on 2WikiMultiHopQA dataset.}
  \label{tab:token_2WikiMultiHopQA}%
\end{table*}%

\begin{table*}[t]
  \centering
    \begin{tabular}{cccccc}
    \toprule
    \multirow{2}[4]{*}{Model} & \multicolumn{5}{c}{\textbf{MusiQue}} \\
\cmidrule{2-6}      & R\&B & R\&L & Ext. & Fil. & E\&F \\
    \midrule
    LongRAG-ChatGLM3-6B-32k w/ SFT & 2141  & 15062  & 2217  & 975  & 1051  \\
    LongRAG-Qwen1.5-7B-32k w/ SFT & 2141  & 15062  & 2198  & 1050  & 1108  \\
    LongRAG-Vicuna-v1.5-7B-16k w/ SFT & 2141  & 14520  & 2240  & 995  & 1094  \\
    LongRAG-Llama3-8B-8k w/ SFT & 2141  & 7711  & 2196  & 828  & 883  \\
    \bottomrule
    \end{tabular}%
    \caption{Values of the token length fed into the generator on MusiQue dataset.}
  \label{tab:token_MusiQue}%
\end{table*}%

\clearpage
\begin{table*}[h]
  \centering
    \begin{tabular}{ccccc}
    \toprule
    \textbf{Datasets} & \textbf{HotpotQA} & \textbf{2WikiMultiHopQA} & \textbf{MusiQue} & \textbf{QASPER} \\
    \midrule
    Num of long-context extractor data & 200 & 200 & 200 & 100 \\
    Num of CoT-guiding data & 200 & 200 & 200 & 100 \\
    Num of filtering data & 200 & 200 & 200 & - \\
    Num of task-oriented data & 200 & 200 & 200 & - \\
    \midrule
    Num of samples & 800 & 800 & 800 & 200 \\
    \bottomrule
    \end{tabular}%
    \caption{Statistics of our fine-tuning instruction dataset \texttt{LRGinstruction}.}
  \label{statistics of our fine-tuning instruction dataset}%
\end{table*}%

\clearpage
\begin{table*}[h]
  \centering
    \begin{tabular}{|p{15cm}|}
    \hline
    \multicolumn{1}{|c|}{\cellcolor{gray!10}\textbf{[STEP-1]}: Data construction prompt for Extractor} \\
    \texttt{\{\textcolor{blue}{supporting paragraphs}\}\newline{}\newline{}
    Based on the above background only, please output the original information that needs to be cited to answer the following questions. Please ensure that the information cited is detailed and comprehensive.\newline{}\newline{}
    Question:\{\textcolor{blue}{question}\}\newline{}\newline{}
    Output only the original information of the required reference: \newline{}
    \{\textcolor{blue}{global information}}\} \\
    \hline
    \multicolumn{1}{|c|}{\cellcolor{gray!10}\textbf{[STEP-2]}: An LLM-based self-evaluator for Extractor} \\
    \texttt{I am going to provide you with a question, the background information, and the answer to that question. Please evaluate whether the answer can be solely derived from the given background information. If it can, set the status value as True, if it can’t, set the status value as False.\newline{}\newline{}
    Question:\{\textcolor{blue}{question}\}\newline{}\newline{}
    Background Information:\{\textcolor{blue}{global information}\}\newline{}\newline{}
    Answer:\{\textcolor{blue}{answer}\}\newline{}\newline{}
    Your output format should be the following json format:\newline{}status: \{the value of status\} }\\
    \hline
    \hline
    \multicolumn{1}{|c|}{\cellcolor{gray!10}\textbf{[RESULT]}: \textbf{Long-Context Extractor Data} for Extractor} \\
    \texttt{\textbf{Instruction}:\newline{}\{\textcolor{blue}{content}\}\newline{}Based on the above background, please output the information you need to cite to answer the question below.\newline{}\{\textcolor{blue}{question}\}\newline{}\newline{}
    \textbf{Output:}\newline{}
    \{\textcolor{blue}{global information}\} }\\
    \hline
    \end{tabular}%
    \caption{Data construction pipeline for extractor and format illustration of long-context extractor data.}
  \label{Data construction prompt for Extractor}%
\end{table*}%

\clearpage
\begin{table*}[h]
  \centering
    \begin{tabular}{|p{15cm}|}
    \hline
    \multicolumn{1}{|c|}{\cellcolor{gray!10}\textbf{[STEP-1]}: Data construction prompt for CoT guidance stage} \\
    \texttt{\{\textcolor{blue}{supporting paragraphs}\}\newline{}\newline{}Given question:\{\textcolor{blue}{question}\}\newline{}\newline{}
    The answer is:\{\textcolor{blue}{answer}\}\newline{}\newline{}
    Your task is to give your thought process for this given question based on the above information, only give me your thought process and do not output other information.\newline{}
    Thought process: \{\textcolor{blue}{CoT}\}
    } \\
    \hline
    \multicolumn{1}{|c|}{\cellcolor{gray!10}\textbf{[STEP-2]}: An LLM-based self-evaluator for CoT guidance stage} \\
    \texttt{Question:\{\textcolor{blue}{question}\}\newline{}\newline{}
    Thought process of the question:\{\textcolor{blue}{CoT}\}\newline{}\newline{}
    Answer:\{\textcolor{blue}{answer}\}\newline{}\newline{}
    Please evaluate whether the thought process of this question can explain the answer to this question. If it can explain the answer, set the value of status to True. If it cannot explain the answer, set the value of status to False.
    Your output format should be the following json format:\newline{}status: \{the value of status\}
    }\\
    \hline
    \hline
    \multicolumn{1}{|c|}{\cellcolor{gray!10}\textbf{[RESULT-1]}: \textbf{CoT-guiding Data} for CoT guidance stage} \\
    \texttt{\textbf{Instruction:}\newline{}\{\textcolor{blue}{content}\}\newline{}
    Please combine the above information and give your thought process for the following \newline{}
    Question:\{\textcolor{blue}{question}\}\newline{}\newline{}
    \textbf{Output:}\newline{}
    \{\textcolor{blue}{CoT}\} }\\
    \hline
    \multicolumn{1}{|c|}{\cellcolor{gray!10}\textbf{[RESULT-2]}: \textbf{Filtering Data} for filtering stage} \\
    \texttt{\textbf{Instruction:}\newline{} Given an article:\{\textcolor{blue}{content}\}\newline{}
    Question:\{\textcolor{blue}{question}\}\newline{}
    Thought process for the question:\{\textcolor{blue}{CoT}\}\newline{}\newline{}
    Your task is to use the thought process provided to decide whether you need to cite the article to answer this question. If you need to cite the article, set the status value to True. If not, set the status value to False. Please output the response in the following json format:\newline{}
    \{"status": \{the value of status\}\}\newline{}\newline{}
    \textbf{Output:}\newline{}
    \{\textcolor{blue}{status}\}}\\
    \hline
    \end{tabular}%
    \caption{Data construction pipeline for filter, and format illustration of CoT-guiding and filtering data.}   
  \label{Data construction prompt for Filter}%
\end{table*}%

\clearpage
\begin{table*}[h]
  \centering
    \begin{tabular}{|p{15cm}|}
    \hline
    \multicolumn{1}{|c|}{\cellcolor{gray!10}\textbf{[RESULT]}: \textbf{Task-Oriented Data} for RAG task} \\
    \texttt{\textbf{Instruction:}\newline{}
    \{\textcolor{blue}{content}\}\newline{}
    Based on the above information, Only give me the answer and do not output any other words.\newline{}
    Question:\{\textcolor{blue}{question}\}\newline{}\newline{}
    \textbf{Output:}\newline{}
    \{\textcolor{blue}{answer}\}}\\
    \hline
    \end{tabular}%
    \caption{Data construction pipeline for RAG task, and format illustration of task-oriented data.}
  \label{Data construction prompt for RAG task}%
\end{table*}%

\clearpage
\begin{table*}[h]
  \centering
  \renewcommand{\arraystretch}{1.5}
    \begin{tabular}{|p{15cm}|}
    \hline
    \cellcolor{blue!10}\textbf{Instruction:}\newline{}\newline{}
    \texttt{Alan Marshal (actor)Alan Marshal( 29 January 1909 – 9 July 1961) was an actor who performed on stage in the United States and in Hollywood films. He was sometimes billed as Alan Marshall or Alan Willey. Hans Tambs LycheHans Tambs Lyche( 21 November 1859 – 16 April 1898) was a Norwegian engineer, unitarian minister, journalist and magazine editor.\newline{}\newline{}
    Alan DeyermondAlan Deyermond FBA( 24 February 1932 – 19 September 2009) was a British professor of Medieval Spanish Literature and Hispanist. His obituary cited him as " the English- speaking world's leading scholar of medieval Hispanic literature. " He spent his academic career associated with one University of London college, Westfield College( later merged with Queen Mary College, to form Queen Mary and Westfield College). \newline{}\newline{}
    Deyermond started his career in 1955 as a lecturer at Westfield College, London. When Westfield merged with Queen Mary College in 1992, he moved to the Mile End site. In the period 1978 – 1980 he held a joint chair at Westfield and at Princeton University. \newline{}\newline{}
    Ralph Tambs-LycheRalph Tambs- Lyche ( 6 September 1890 – 15 January 1991) was a Norwegian mathematician. He was born in Macon, Georgia as a son of Norwegian father Hans Tambs Lyche( 1859 – 1898) and American mother Mary Rebecca Godden( 1856 – 1938). He moved to Norway at the age of two.\newline{}
    \newline{}\ldots\ldots\newline{}\newline{}
    Alan M. WaldAlan Maynard Wald, usually Alan M. Wald or Alan Wald, is an American professor emeritus of English Literature and American Culture at the University of Michigan, Ann Arbor, and writer of 20th-Century American literature who focuses on Communist writers; he is an expert on the American 20th-Century  "Literary Left. " Allen SangreeAllen Luther Sangree, also as Allan or Alan( c. 1878 – March 2, 1924) was an American sports writer and war journalist. \newline{}\newline{}
    \textbf{Based on the above background, please output the information you need to cite to answer the question below.} \newline{}
    Who died earlier, Hans Tambs Lyche or Alan Deyermond?}\newline{} \\
    \hline
    \hline
    \cellcolor{yellow!15}\textbf{Output:}\newline{}\newline{}
    \texttt{Hans Tambs Lyche (21 November 1859 – 16 April 1898) was a Norwegian engineer, unitarian minister, journalist, and magazine editor.  Alan Deyermond (24 February 1932 – 19 September 2009) was a British professor of Medieval Spanish Literature and Hispanist.\newline{}} \\
    \hline
    \end{tabular}%
    \caption{An example of long-context extractor data of \texttt{LRGinstruction}.}
  \label{tab:construction dataset: cite}%
\end{table*}%

\clearpage
\begin{table*}[h]
  \centering
\renewcommand{\arraystretch}{1.5}
    \begin{tabular}{|p{15cm}|}
    \hline
    \cellcolor{blue!10}\textbf{Instruction:}\newline{}\newline{}
    \texttt{Beatrice of Castile, Marchioness of MontferratBeatrice of Castile (5 November or 6 December 1254 - 1286) was a daughter of Alfonso X of Castile and his wife Violant of Aragon. She was Marchioness of Montferrat by her marriage and was mother of Irene of Montferrat. \newline{}\newline{}
    Theodred II (Bishop of Elmham)Theodred II was a medieval Bishop of Elmham. The date of Theodred's consecration unknown, but the date of his death was sometime between 995 and 997. Irene of MontferratYolande of Montferrat (– 1317 in Constantinople) (also known as Violante, then Empress Irene) was the second wife of Andronikos II Palaiologos and thus empress of the Byzantine Empire. She was the heir of the Margraviate of Montferrat. Born in Casale, she was daughter of William VII, Marquess of Montferrat and his second wife Beatrice of Castile. Her maternal grandparents were King Alfonso X of Castile and his wife Violante of Aragon. Yolande (variation of Violante) was named after her grandmother.\newline{}\newline{}
    In 1284, Andronikos II, a widower by his first marriage with Anna of Hungary, married Yolanda (who was renamed Eirene as Empress). She and Andronikos II were distant cousins, both being descendants of Andronikos Doukas Angelos (ca. 1122-1185). With her, Eirene brought the Montferrat rights to the kingdom of Thessalonica, a dominion that, despite having been conquered half-a-century before Eirene's birth by the Byzantine state of Epirus, was still claimed by its short-lived (1204–1224) Montferrat royal dynasty.\newline{}
    \newline{}\ldots\ldots\newline{}\newline{}
    Henry, Marquess of MontferratHenry( c.1020 -c.1044/5) was a member of the Aleramid dynasty. He was a younger son of William III of Montferrat and Waza. From 1042, he was co-ruler of the March of Montferrat with his older brother Otto II of Montferrat. Probably in 1041, certainly before 19 January 1042, he married Adelaide of Susa, the heiress of the March of Turin, which temporarily united the two great northwestern Italian marches of Turin and Montferrat. \newline{} \newline{}
    \textbf{Please combine the above information and give your thought process for the following}\newline{}
    \textbf{Question:}  What is the date of death of Irene Of Montferrat's father?}\newline{} \\
    \hline
    \hline
    \cellcolor{yellow!15}\textbf{Output:} \newline{}\newline{}
    \texttt{To find the date of death of Irene Of Montferrat's father, we can look at the information provided about William VII, Marquess of Montferrat. It states that he died on 6 February 1292. Since Irene was born around 1253, William VII would have been her father. Therefore, the date of death of Irene Of Montferrat's father is 6 February 1292.\newline{}} \\
    \hline
    \end{tabular}%
    \caption{An example of CoT-guiding data of \texttt{LRGinstruction}.}
  \label{tab:construction dataset: CoT}%
\end{table*}%

\clearpage
\begin{table*}[h]
  \centering
  \renewcommand{\arraystretch}{1.5}
    \begin{tabular}{|p{15cm}|}
    \hline
    \cellcolor{blue!10}\textbf{Instruction:}\newline{}\newline{}
    \texttt{\textbf{Given an article:}"Awaiting on You All" is a song by English musician George Harrison, released on his 1970 triple album, "All Things Must Pass". Along with the single "My Sweet Lord", it is among the more overtly religious compositions on "All Things Must Pass", and the recording typifies co-producer Phil Spector's influence on the album, due to his liberal use of reverberation and other Wall of Sound production techniques. \newline{}\newline{}
    Harrison recorded the track in London backed by musicians such as Eric Clapton, Bobby Whitlock, Klaus Voormann, Jim Gordon and Jim Price – many of whom he had toured with, as Delaney \& Bonnie and Friends, in December 1969, while still officially a member of the Beatles. Musically, the composition reflects Harrison's embracing of the gospel music genre, following his production of fellow Apple Records artists Billy Preston and Doris Troy.\newline{}
    \newline{}\ldots\ldots\newline{}\newline{}
    A similarly well-regarded live version, with backing from a large band including Clapton, Ringo Starr, Preston and Jim Keltner, was released on the 1971 album "The Concert for Bangladesh" and appeared in the 1972 film of the same name. Harrison's posthumous compilation (2012) includes a demo version of the song, recorded early in the 1970 sessions for "All Things Must Pass".\newline{}\newline{}
    \textbf{Question:} What is the date of death of the performer of song Awaiting On You All?\newline{}\newline{}
    \textbf{Thought process for the question:}The question asks for the date of death of the performer of the song "Awaiting on You All." We know from the given information that the song was written and performed by English musician George Harrison. To find his date of death, we can look for the date of death of George Harrison in the text. We find that George Harrison died on 29 November 2001. Therefore, the answer to the question is 29 November 2001.\newline{}\newline{}
    \textbf{Your task is to use the thought process provided to decide whether you need to cite the article to answer this question. If you need to cite the article, set the status value to True. If not, set the status value to False. Please output the response in the following json format:\newline{}
    \{"status": \{the value of status\}\}\newline{}}} \\
    \hline
    \hline
    \cellcolor{yellow!15}\textbf{Output:} \newline{}\newline{}
    \texttt{\{"status": \{"True"\}\} \newline{}}\\
    \hline
    \end{tabular}%
    \caption{An example of filtering data of \texttt{LRGinstruction}.}
  \label{tab:construction dataset: Filter}%
\end{table*}%

\clearpage
\begin{table*}[h]
  \centering
  \renewcommand{\arraystretch}{1.5}
    \begin{tabular}{|p{15cm}|}
    \hline
    \cellcolor{blue!10}\textbf{Instruction:}\newline{}\newline{}
    \texttt{My Name Is Anthony Gonsalves (film) My Name Is Anthony Gonsalves is a Bollywood drama film starring newcomer Nikhil Dwivedi, Amrita Rao and Mithun Chakraborty as the lead protagonists. The film is directed by Eeshwar Nivas. The name of the movie is derived from the 1977 hit movie Amar Akbar Anthony's famous song," My Name Is Anthony Gonsalves." It was released on 11 January 2008 and was a box office bomb.\newline{}\newline{}
    My Name Is JuaniMy Name Is Juani is a 2006 Spanish drama film written and directed by Bigas Luna. My Name Is BanduMy Name is Bandu is a 2015 Sri Lankan Sinhala comedy, family film directed by Suranga de Alwis and produced by Suranga de Alwis. It stars Bandu Samarasinghe, and Anusha Damayanthi in lead roles along with Rodney Warnakula, Roy de Silva and Mark Samson. Music for the film is done by Sarath de Alwis. The film is the 85th film of Bandu Samarasinghe. It is the 1239th Sri Lankan film in the Sinhala cinema.\newline{}\newline{}
    My Name Is KhanMy Name Is Khan is a 2010 Indian Hindi- language drama film directed by Karan Johar, produced by Hiroo Johar and Gauri Khan, and starring Shah Rukh Khan and Kajol in lead roles.\newline{}
    \newline{}\ldots\ldots\newline{}\newline{}
    The film stars Shakib Khan and Sahara in the lead roles, with Ahmed Sharif, Misha Shoudagor, Probir Mitro and Rahena Joli playing other significant roles in the film. \newline{}\newline{}
    My Name Is Sultan was released on 20 August 2012. Leslie, My Name Is EvilLeslie, My Name Is Evil is a 2009 Canadian film written and directed by Reginald Harkema. It was renamed" Manson, My Name Is Evil" after its initial release.\newline{}\newline{}
    My Name Is NobodyMy Name Is Nobody is a 1973 comedy spaghetti western starring Terence Hill and Henry Fonda. The film was directed by Tonino Valerii.\newline{}\newline{}
    My Name Is Rocco PapaleoMy Name Is Rocco Papaleo is a 1971 Italian comedy film directed by Ettore Scola.  \newline{} \newline{}
    \textbf{Based on the above information, Only give me the answer and do not output any other words.}\newline{}
    \textbf{Question: }Which film was released more recently, My Name Is Bandu or Leadbelly (Film)? \newline{}
    \textbf{Answer:}\newline{}} \\
    \hline
    \hline
    \cellcolor{yellow!15}\textbf{Output:} \newline{}\newline{}
    \texttt{My Name Is Bandu\newline{}}\\
    \hline
    \end{tabular}%
    \caption{An example of task-oriented data of \texttt{LRGinstruction}.}
  \label{tab:construction dataset: RAG task}%
\end{table*}%

\clearpage
\begin{table*}[h]
  \centering
    \begin{tabular}{|p{15cm}|}
    \hline
    \multicolumn{1}{|c|}{\cellcolor{gray!10}\textbf{Prompt of LLM-augmented information extractor}}
    \\    \texttt{\textbf{Instruction:}\newline{}\{\textcolor{blue}{content}\}\newline{}Based on the above background, please output the information you need to cite to answer the question below.\newline{}\{\textcolor{blue}{question}\}\newline{}\newline{}
    \textbf{Output:}\newline{}
    \{\textcolor{blue}{global information}\} }\\
    \hline
    \multicolumn{1}{|c|}{\cellcolor{gray!10}\textbf{Prompt of CoT guidance stage in CoT-guided filter}} \\
    \texttt{\textbf{Instruction:}\newline{}\{\textcolor{blue}{content}\}\newline{}
    Please combine the above information and give your thought process for the following\newline{}
    Question:\{\textcolor{blue}{question}\}\newline{}\newline{}
    \textbf{Output:}\newline{}
    \{\textcolor{blue}{CoT}\} }\\
    \hline
    \multicolumn{1}{|c|}{\cellcolor{gray!10}\textbf{Prompt of filtering stage in CoT-guided filter}} \\
    \texttt{\textbf{Instruction:}\newline{} Given an article:\{\textcolor{blue}{content}\}\newline{}
    Question:\{\textcolor{blue}{question}\}\newline{}
    Thought process for the question:\{\textcolor{blue}{CoT}\}\newline{}\newline{}
    Your task is to use the thought process provided to decide whether you need to cite the article to answer this question. If you need to cite the article, set the status value to True. If not, set the status value to False. Please output the response in the following json format:\newline{}
    \{"status": \{the value of status\}\}\newline{}\newline{}
    \textbf{Output:}\newline{}
    \{\textcolor{blue}{status}\}}\\
    \multicolumn{1}{|c|}{\cellcolor{gray!10}\textbf{Prompt of LLM-augmented generator}} \\
    \texttt{\textbf{Instruction:}\newline{}
    \{\textcolor{blue}{content}\}\newline{}
    Based on the above information, Only give me the answer and do not output any other words.\newline{}
    Question:\{\textcolor{blue}{question}\}\newline{}\newline{}
    \textbf{Output:}\newline{}
    \{\textcolor{blue}{answer}\}}\\
    \hline
    \end{tabular}%
    \caption{All prompts of LongRAG system.}
  \label{tab:Prompts of LongRAG system}%
\end{table*}%

\clearpage
\begin{table*}[h]
  \centering
    \begin{tabular}{l}
    \toprule
    \textbf{Question: Where did the performer of song I'll Say It graduate from?} \\
    \midrule
    \multicolumn{1}{p{40em}}{
    \textbf{Input to generator (2082 tokens):} \newline{}
    Answer the question based on the given passages. Only give me the answer and do not output any other words. The following are given passages.  The duo promoted the song by performing it on various television shows and at various venues, of which included GMTV and Sony Ericsson  's Dance Nation Festival.  This was planned to be the first single off the band  's second studio album Say It Now, which was scheduled for release in November 2009, but due to the low chart placing of "Say It", the album was eventually cancelled.   Background "Say It" was written by Carl Björsell, Didrik Thott and Sebastian Thott.
    \newline{}\ldots\ldots\newline{}
    We just want to show progression."The song was composed in a key of C sharp minor and runs at a tempo of 126.96 beats per minute. The song was produced with consistence of various drum and bass and electronica instrumentation.Passage 1:
    \textcolor{blue}{ \hlgreen{I  'll Say It} "I  'll Say It" is a  \hlgreen{song} written by American musician Adam Schlesinger and  \hlgreen{recorded by} comedian Kathy  \hlgreen{Griffin}, released as the theme song for her show, Kathy.}
    It was additionally used as the introduction music to her 2012 comedy special "Kennedie Center on Hers" and continued to be used in future specials.
    \textcolor{blue}{On August 20, 2012, Griffin released a seven track EP containing dance remixes of "I  'll Say It".}
    Music video  The music video begins in the day with Kathy Griffin in her house preparing her make-up. It shows her daily routine visiting her dogs, leaving the house and driving to a theater, ending with her on stage in her signature pose. The scenes are interlaced with various clips of Los Angeles, California.Passage 10:  Say It (Booty Luv song) "Say It" is a song by female English dance music duo Booty Luv.
    \newline{}\ldots\ldots\newline{}
    Filmography  Film  Television  Other  Stand-up specials  Discography  On June 10, 2008, Griffin released a comedy CD titled For Your Consideration. The disc was recorded at the ETK Theatre at the Grand Theatre Center For The Arts in Tracy, California on February 17, 2008. Griffin stated she decided to release the CD to try to win a Grammy award.On August 25, 2009, Griffin released a second comedy album, Suckin  ' It for the Holidays, in another bid for a Grammy.
    \textcolor{blue}{Griffin received her third Grammy nomination for Kathy Griffin: Does the Bible Belt in 2010,.On May 4, 2012, the full length version of "I  'll Say It", the theme song of her show Kathy, was released to iTunes as a single.  On August 20, 2012, Griffin released a seven-track EP containing dance remixes of "I  'll Say It".}
    Bibliography  Official Book Club Selection: A Memoir According to Kathy Griffin. Ballantine Books. 2009. ISBN 978-0345518569.  Kathy Griffin  's Celebrity Run-Ins: My A-Z Index. Flatiron Books. 2016. ISBN 978-1250115638. Song went on a five-year hiatus from acting. She became an adjunct professor and part-time lecturer at \hlred{Seoul Arts College} in 2010, as a faculty member of the Department of Performing Arts and the Department of Broadcasting, Entertainment and Visual Arts.
    \newline{}\ldots\ldots\newline{}
    Asher Roth sampled the song for his debut rap single "I Love College". After the song leaked onto the internet, Rivers Cuomo reportedly refused to clear the sample, which prompted Roth to debut a remixed version of his song as his official debut single.   Answer the question based on the given passages. Only give me the answer and do not output any other words.   \newline{}\newline{}
    Question: Where did the performer of song I  'Ll Say It graduate from?  
    \newline{}Answer:} \\
    \midrule
    \textbf{Answer of RAG-base}: \hlred{Seoul Arts College} \textcolor{red}{\ding{55}} \\
    \textbf{Golden Answer}: \hlgreen{Lee Strasberg Theatre and Film Institute} \textcolor{green}{\ding{51}} \\
    \midrule
    \textbf{Wrong Reason}: \textcolor{red}{Incomplete key information} \\
    \bottomrule
    \end{tabular}%
    \caption{A question-answering example of Vanilla RAG (RAG-Base). \hlgreen{The words in the green area} indicate correct relevant information and answers while \hlred{red} means the opposite. \textcolor{blue}{The blue snippets} are question-relevant information. The correct answer is labeled "\textcolor{green}{\ding{51}}", while wrong answer labeled "\textcolor{red}{\ding{55}}".}
  \label{tab:System Case: RAG-base}%
\end{table*}%

\clearpage
\begin{table*}[h]
  \centering
    \begin{tabular}{l}
    \toprule
    \textbf{Question: Where did the performer of song I'Ll Say It graduate from?} \\
    \midrule
    \multicolumn{1}{p{40em}}{
    \textbf{Input to generator (23047 tokens):} 
    \newline{}
    Answer the question based on the given passages. Only give me the answer and do not output any other words.The following are given passages.
    \newline{}\ldots\ldots\newline{}
    The girls then head downstairs to a mini casino where they gamble. The girls are then seen against various backgrounds and laying on chairs. Finally, the girls have a party in their hotel room and invite their friends and some men to their hotel rooms, before sending them away. Chart performance Weekly charts Year-end charts Passage 1:
    \textcolor{blue}{\hlgreen{I'll Say It}"I'll Say It" is a  \hlgreen{song} written by American musician Adam Schlesinger and  \hlgreen{recorded by} comedian Kathy \hlgreen{Griffin}, released as the theme song for her show, Kathy.}
    It was additionally used as the introduction music to her 2012 comedy special "Kennedie Center on Hers" and continued to be used in future specials.
    \textcolor{blue}{On August 20, 2012, Griffin released a seven track EP containing dance remixes of "I  'll Say It".}
    Music video The music video begins in the day with Kathy Griffin in her house preparing her make-up. It shows her daily routine visiting her dogs, leaving the house and driving to a theater, ending with her on stage in her signature pose. The scenes are interlaced with various clips of Los Angeles, California.Charts Passage 2:Kathy Griffin Kathleen Mary Griffin (born November 4, 1960) is an American comedian and actress who has starred in television comedy specials and has released comedy albums. In 2007 and 2008, Griffin won Primetime Emmy Awards for her reality show Kathy Griffin: My Life on the D-List. She has also appeared in supporting roles in films.
    \textcolor{blue}{\hlgreen{Griffin} was born in Oak Park, Illinois. In 1978, she moved to Los Angeles, where she \hlgreen{studied drama at the Lee Strasberg Theatre and Film Institute} and became a member of the improvisational comedy troupe The Groundlings.}
    In the 1990s, Griffin began performing as a stand-up comedian and appeared as a guest star on television shows, including a supporting role on the NBC sitcom Suddenly Susan (1996–2000).
    \newline{}\ldots\ldots\newline{}
    \textcolor{blue}{Griffin released a second comedy album, Suckin' It for the Holidays, in another bid for a Grammy.Griffin received her third Grammy nomination for Kathy Griffin: Does the Bible Belt in 2010,.On May 4, 2012, the full length version of "I'll Say It", the theme song of her show Kathy, was released to iTunes as a single.On August 20, 2012, Griffin released a seven-track EP containing dance remixes of "I'll Say It".}
    \newline{}\ldots\ldots\newline{}
    Song Yoon-ah was born in Seoul, but spent her childhood in Gimcheon, North Gyeongsang Province. She has two elder brothers, the first one is a doctor. While studying Cultural Anthropology as a freshman at \hlred{Hanyang University}, she was recommended by an older schoolmate to a modeling agency.
    \newline{}\ldots\ldots\newline{}
     Chiptune artist Inverse Phase parodied the song on a Commodore 64, titling it "Say It Ain't Sixty-FO" Calpurnia covered the song for Spotify's Under Cover podcast in 2018 In popular culture "Say It Ain't So" is a playable track in the video games Rock Band and Rocksmith 2014 in addition to appearing on an episode of Hindsight. Answer the question based on the given passages. Only give me the answer and do not output any other words. \newline{}\newline{}
    Question: Where did the performer of song I'll Say It graduate from?
    \newline{}
    Answer:} \\
    \midrule
    \textbf{Answer of RAG-Long}: \hlred{Hanyang University} \textcolor{red}{\ding{55}} \\
    \textbf{Golden Answer}: \hlgreen{Lee Strasberg Theatre and Film Institute} \textcolor{green}{\ding{51}} \\
    \midrule
    \textbf{Wrong Reason}: \textcolor{red}{Complete key information but lost in middle} \\
    \bottomrule
    \end{tabular}%
    \caption{A question-answering example of our LongRAG with RAG-Long component strategy. \hlgreen{The words in the green area} indicate correct relevant information and answers while \hlred{red} means the opposite. \textcolor{blue}{The blue snippets} are question-relevant information. The correct answer is labeled "\textcolor{green}{\ding{51}}", while wrong answer labeled "\textcolor{red}{\ding{55}}".}
  \label{tab:System Case: RAG-Long}%
\end{table*}%

\clearpage
\begin{table*}[h]
  \centering
    \begin{tabular}{l}
    \toprule
    \textbf{Question: Where did the performer of song I'Ll Say It graduate from?} \\
    \midrule
    \multicolumn{1}{p{40em}}{
    \textbf{Input to generator (644 tokens):} 
    \newline{}
    Answer the question based on the given passages. Only give me the answer and do not output any other words.The following are given passages.Passage 1:
    \textcolor{blue}{ \hlgreen{I'll Say It}"I'll Say It" is a  \hlgreen{song} written by American musician Adam Schlesinger and  \hlgreen{recorded by} comedian Kathy  \hlgreen{Griffin}, released as the theme song for her show, Kathy.}
    It was additionally used as the introduction music to her 2012 comedy special "Kennedie Center on Hers" and continued to be used in future specials.
    \textcolor{blue}{On August 20, 2012, Griffin released a seven track EP containing dance remixes of "I'll Say It".}
    Music video The music video begins in the day with Kathy Griffin in her house preparing her make-up. It shows her daily routine visiting her dogs, leaving the house and driving to a theater, ending with her on stage in her signature pose. The scenes are interlaced with various clips of Los Angeles, California. in a ceremony officiated by comedian Lily Tomlin. Filmography Film Television Other Stand-up specials Discography On June 10, 2008, Griffin released a comedy CD titled For Your Consideration. The disc was recorded at the ETK Theatre at the Grand Theatre Center For The Arts in Tracy, California on February 17, 2008. Griffin stated she decided to release the CD to try to win a Grammy award. On August 25, 2009, Griffin released a second comedy album, Suckin' It for the Holidays, in another bid for a Grammy. Griffin received her third Grammy nomination for Kathy Griffin: Does the Bible Belt in 2010,.On May 4, 2012, the full length version of "I'll Say It", the theme song of her show Kathy, was released to iTunes as a single. On August 20, 2012, Griffin released a seven-track EP containing dance remixes of "I'll Say It".  Bibliography Official Book Club Selection: A Memoir According to Kathy Griffin. Ballantine Books. 2009. ISBN 978-0345518569. Kathy Griffin's Celebrity Run-Ins: My A-Z Index. Flatiron Books. 2016. ISBN 978-1250115638. 
    \textcolor{blue}{The  \hlgreen{performer of the song "I'll Say It"} is Kathy  \hlgreen{Griffin}, an American comedian and actress who has starred in television comedy specials and has released comedy albums. She attended the  \hlgreen{Lee Strasberg Theatre and Film Institute} in Los Angeles, where she  \hlgreen{studied drama}.}
    Answer the question based on the given passages. Only give me the answer and do not output any other words.\newline{}  Question: Where did the performer of song I'll Say It graduate from? \newline{}
    Answer:} \\
    \midrule
    \textbf{Answer of LongRAG}: \hlgreen{Lee Strasberg Theatre and Film Institute} \textcolor{green}{\ding{51}} \\
    \textbf{Golden Answer}: \hlgreen{Lee Strasberg Theatre and Film Institute} \textcolor{green}{\ding{51}} \\
    \bottomrule
    \end{tabular}%
    \caption{A question-answering example of our LongRAG system with E\&F component strategy. \hlgreen{The words in the green area} indicate correct relevant information and answers while \hlred{red} means the opposite. \textcolor{blue}{The blue snippets} are question-relevant information. The correct answer is labeled "\textcolor{green}{\ding{51}}", while wrong answer labeled "\textcolor{red}{\ding{55}}".}
  \label{tab:System Case: Extractor and Filter}%
\end{table*}%

\end{document}